\documentclass[10pt,twocolumn,letterpaper]{article}

\usepackage{slashbox}
\usepackage[pagenumbers]{cvpr} % To force page numbers, \eg for an arXiv version
\makeatletter
\@namedef{ver@everyshi.sty}{}
\makeatother
\usepackage{tikz}
\def\halfcheckmark{\tikz\draw[scale=0.4,fill=black](0,.35) -- (.25,0) -- (1,.7) -- (.25,.15) -- cycle (0.75,0.2) -- (0.77,0.2)  -- (0.6,0.7) -- cycle;}

\usepackage{graphicx}
\usepackage{amsmath}
\usepackage{amssymb}
\usepackage{booktabs}
\usepackage{placeins}
\usepackage[normalem]{ulem}
\usepackage{multirow}
\usepackage{pifont}
\newcommand{\cmark}{\ding{51}}%
\newcommand{\xmark}{\ding{55}}%
\usepackage{subcaption,siunitx,booktabs}
\usepackage[font={small}]{caption}
\usepackage[misc]{ifsym}

\usepackage[pagebackref,breaklinks,colorlinks]{hyperref}

\usepackage[capitalize]{cleveref}
\crefname{section}{Section}{Secs.}
\Crefname{section}{Section}{Sections}
\Crefname{table}{Table}{Tables}
\crefname{table}{Tab.}{Tabs.}

\def\cvprPaperID{2379} % *** Enter the CVPR Paper ID here
\def\confName{CVPR}
\def\confYear{2023}

\begin{document}

\title{\vspace{-0.4cm}CelebV-Text: A Large-Scale Facial Text-Video Dataset\vspace{-0.4cm}}

\renewcommand*{\thefootnote}{\fnsymbol{footnote}}

% AUTHOR LIST
\author{
Jianhui Yu$^{1*}$ \quad Hao Zhu$^{2*}$ \quad Liming Jiang$^3$ \quad Chen Change Loy$^3$ \quad Weidong Cai$^1$ \quad Wayne Wu$^{4}$\\
{\small $^1$University of Sydney \quad $^2$SenseTime Research \quad $^3$S-Lab, Nanyang Technological University \quad $^4$Shanghai AI Laboratory}\\
{\tt\small 
jianhui.yu@sydney.edu.au \quad 
haozhu96@gmail.com  \quad 
\{liming002,ccloy\}@ntu.edu.sg
} \\
\vspace{-3mm}
{\tt\small 
tom.cai@sydney.edu.au \quad 
wuwenyan0503@gmail.com}
}

\twocolumn[{%
\renewcommand\twocolumn[1][]{#1}%
\maketitle
\begin{center}
  % \captionsetup{type=figure}
  \includegraphics[width=1.0\linewidth]{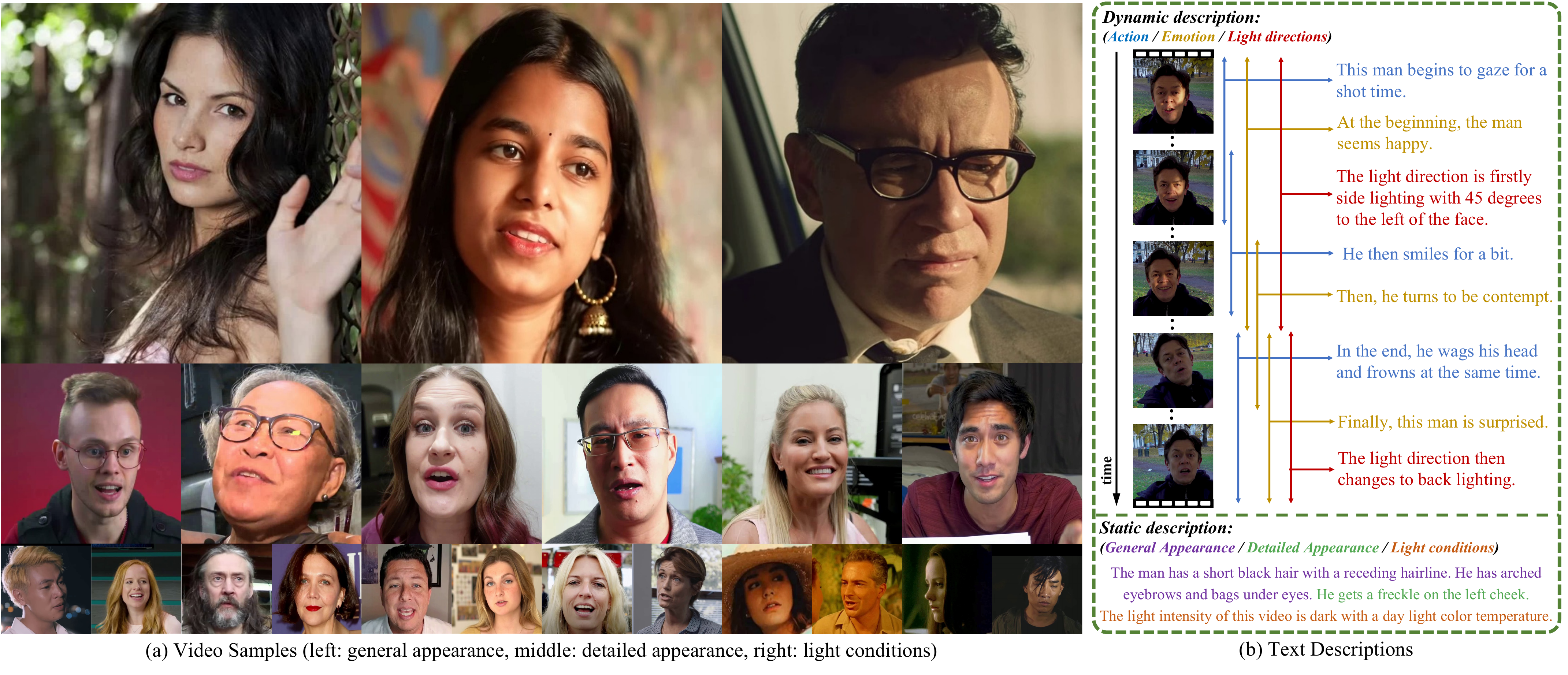}
  \vspace{-5mm}
  \captionof{figure}{\textbf{Overview of CelebV-Text.} CelebV-Text contains (a) $70,000$ video samples and (b) $1,400,000$ text descriptions. Each video sample is annotated with general appearance, detailed appearance, light conditions, action, emotion, and light directions.}
  \label{fig:teaser}
\end{center}%
}]

\def\thefootnote{*}\footnotetext{Equal contribution.}
\def\thefootnote{\arabic{footnote}}
% \def\thefootnote{\Letter}\footnotetext{Corresponding author.}
% \def\thefootnote{\arabic{footnote}}

%%%%%%%%% ABSTRACT
\begin{abstract}
% \vspace{-2mm}
\if 0
Currently, text-driven generation models are booming in video editing with their compelling results.
However, for the face-centric text-to-video generation, challenges remain severe as a suitable dataset with high-quality videos and highly-relevant texts is lacking.
In this work, we present a large-scale, high-quality, and diverse facial text-video dataset, \textbf{CelebV-Text}, to facilitate the research of facial text-to-video generation tasks.
%
CelebV-Text contains $70,000$ in-the-wild face video clips covering diverse visual content.
Each video clip is paired with $20$ texts generated by the proposed semi-auto text generation strategy, which is able to describe both the static and dynamic attributes precisely.
We make comprehensive statistical analysis on videos, texts, and text-video relevance of CelebV-Text, verifying its superiority over other datasets.
%
Also, we conduct extensive self-evaluations to show the effectiveness and potential of CelebV-Text.
Furthermore, a benchmark is constructed with representative methods to standardize the evaluation of the facial text-to-video generation task. 
All the data and models are publicly available\footnote{Project page: \url{https://celebv-text.github.io}}.
\fi

Text-driven generation models are flourishing in video generation and editing. However, face-centric text-to-video generation remains a challenge due to the lack of a suitable dataset containing high-quality videos and highly relevant texts. This paper presents \textbf{CelebV-Text}, a large-scale, diverse, and high-quality dataset of facial text-video pairs, to facilitate research on facial text-to-video generation tasks. CelebV-Text comprises 70,000 in-the-wild face video clips with diverse visual content, each paired with 20 texts generated using the proposed semi-automatic text generation strategy. The provided texts are of high quality, describing both static and dynamic attributes precisely. 
The superiority of CelebV-Text over other datasets is demonstrated via comprehensive statistical analysis of the videos, texts, and text-video relevance. The effectiveness and potential of CelebV-Text are further shown through extensive self-evaluation. A benchmark is constructed with representative methods to standardize the evaluation of the facial text-to-video generation task. All data and models are publicly available\footnote{Project page: \url{https://celebv-text.github.io}}.

\end{abstract}
%%%%%%%%% BODY TEXT
\section{Introduction}

\if 0
Recently, text-driven video generation has attracted large attention in computer vision and computer graphics.
Using simple text as input, video motion can be freely generated and controlled, which motivates a large number of applications in both academia and industry~\cite{v2t, tfgan, ranzato2014video, kth}.
However, current development in text-to-video generation meets severe problems, especially in the \textit{face-centric} scenario, where generated video frames are often of low quality~\cite{gupta2018imagine, liu2019cross, v2t} or weakly relevant to input texts~\cite{frozen,crosstask,malmaud2015s,alayrac2016unsupervised}.
We argue that one of the key problems is the lack of a well-suited facial text-video dataset, where high-quality video samples and text descriptions of various attributes highly relevant to videos should be included.
% due to the dataset scale and difficulty in collecting paired text-video datasets~\cite{phenaki, imagen, imagen-video, makeavideo, tats}.
% To replicate this success from image to video domain, there is an urgent need for a well-suited dataset.
% \wayne{1. directly start from text-driven video generation. a) text-driven video editing is important in many domains b) the situation of text-driven video editing: still with poor quality and controllability c) lack of a dataset. (about 10 lines)}
\fi

Text-driven video generation has recently garnered significant attention in the fields of computer vision and computer graphics. By using text as input, video content can be generated and controlled, inspiring numerous applications in both academia and industry \cite{v2t, tfgan, ranzato2014video, kth}.
However, text-to-video generation still faces many challenges, particularly in the face-centric scenario where generated video frames often lack quality \cite{gupta2018imagine, liu2019cross, v2t} or have weak relevance to input texts \cite{frozen, crosstask, malmaud2015s, alayrac2016unsupervised}.
We believe that one of the main issues is the absence of a well-suited facial text-video dataset containing high-quality video samples and text descriptions of various attributes highly relevant to videos.

% 提出建立数据集的挑战 (2) -> 对应的挑战的解决方法 (3.1)
% 直接给标号: 1,2,3
% Challenges of text-to-video generation include not only achieving high fidelity and natural transitions, but also content relevance between texts and videos in terms of dynamic and static aspects, meaning texts need to describe attributes that might or might not change over time.
% However, existing publicly available text-video datasets either synthesized with limited content~\cite{kth, bouncing-mnist, gupta2018imagine,tfgan} or collected in-the-wild with weakly relevant text-video pairs without describing temporal information~\cite{howto100m, frozen,crosstask,malmaud2015s,alayrac2016unsupervised}.
% In addition, there are few face videos to enable face-specific video generation tasks~\cite{celebv, voxceleb, voxceleb2, celebvhq, faceforensics}.
% Currently, the only available text-to-face video dataset is based on the annotation of the Vox dataset~\cite{mmvid}, where text descriptions is generated from fixed appearance labels. 

\if
The construction of such a well-suited facial text-video dataset is challenging mainly in three aspects:
% zhuhao:
% 1. video quantity and quality -> (Section 3.1) 
% 2. text relevance -> (Section 3.2)
% 3. diversity and natualness of text -> (Section 3.3)
% 1) As in most face video datasets~\cite{celebv, voxceleb, voxceleb2, faceforensics}, the quality and quantity of collected video samples largely determine the model performance for video generation~\misscite. Moreover, as videos present abundant spatial and temporal information, collecting diverse videos in a large scale and high quality is quite important. [?]
% 2) Quality and quantity of attribute annotations are also important for generation tasks~\cite{celebvhq}, as they are utilized for text generation that would greatly affect the relevance of text-video pairs. 
% \zhuhao{
1) \textit{Data collection.} The quality and quantity of video samples are important since they largely determine the quality of generated videos~\cite{celebv, voxceleb, voxceleb2, faceforensics}.
However, since the overall distribution of data should be natural and the motion in videos has to be smooth, it is difficult to collect such a large-scale dataset with high-quality.
 % (Section 3.1) 
% 2) The relevance of text-video pairs also needs to be guaranteed. Achieving flexible control of facial text-to-video generation has high requirements for this property. % (Section 3.2) 
2) \textit{Data annotation.} The relevance of text-video pairs needs to be ensured. It requires high coverage of text for the description of content and motion appearing in the video, such as light conditions and head movements. % (Section 3.2) 
3) \textit{Text generation.} Producing diverse, natural, and scalable texts are non-trivial. Manual-text generation is costly and not scalable, while auto-text generation is easily extensible but has limited naturalness. Thus, we need a text generation strategy, which is both scalable and natural.
% (sec. 3.3)
% the difficutlty comes from the disadvantages of both mannual texts and automatic text generation
\fi

Constructing a high-quality facial text-video dataset poses several challenges, mainly in three aspects. 
1) \textit{Data collection.}
The quality and quantity of video samples largely determine the quality of generated videos \cite{celebv, voxceleb, voxceleb2, faceforensics}. However, obtaining such a large-scale dataset with high-quality samples while maintaining a natural distribution and smooth video motion is challenging. 
2) \textit{Data annotation.}
The relevance of text-video pairs needs to be ensured.
This requires a comprehensive coverage of text for describing the content and motion appearing in the video, such as light conditions and head movements. 
3) \textit{Text generation.}
Producing diverse and natural texts are non-trivial. Manual text generation is expensive and not scalable. While auto-text generation is easily extensible, it is limited in naturalness.

To overcome the challenges mentioned above, we carefully design a comprehensive data construction pipeline that includes data collection and processing, data annotation, and semi-auto text generation. 
First, to obtain raw videos, we follow the data collection steps of CelebV-HQ, which has proven to be effective in \cite{celebvhq}. 
We introduce a minor modification to the video processing step to improve the video's smoothness further. Next, to ensure highly relevant text-video pairs, we analyze videos from both temporal dynamics and static content and establish a set of attributes that may or may not change over time.
Finally, we propose a semi-auto template-based method to generate texts that are diverse and natural. Our approach leverages the advantages of both auto- and manual-text methods. Specifically, we design a rich variety of grammar templates as \cite{nlpparser, stap2020conditional} to parse annotation and manual texts, which are flexibly combined and modified to achieve high diversity, complexity, and naturalness.

With the proposed pipeline, we create \textbf{CelebV-Text}, a Large-Scale Facial Text-Video Dataset, which includes $70,000$ in-the-wild video clips with a resolution of at least $512\times512$ and $1,400,000$ text descriptions with 20 for each clip. 
As depicted in Figure~\ref{fig:teaser}, CelebV-Text consists of high-quality video samples and text descriptions for realistic face video generation. 
Each video is annotated with three types of static attributes (40 general appearances, 5 detailed appearances, and 6 light conditions) and three types of dynamic attributes (37 actions, 8 emotions, and 6 light directions). All dynamic attributes are densely annotated with start and end timestamps, while manual-texts are provided for labels that cannot be discretized. Furthermore, we have designed three templates for each attribute type, resulting in a total of 18 templates that can be flexibly combined. All attributes and manual-texts are naturally described in our generated texts.

CelebV-Text surpasses existing face video datasets~\cite{voxceleb2} in terms of resolution (over 2 times higher), number of samples, and more diverse distribution. In addition, the texts in CelebV-Text exhibit higher diversity, complexity, and naturalness than those in text-video datasets~\cite{mmvid, celebvhq}. 
CelebV-Text also shows high relevance of text-video pairs, validated by our text-video retrieval experiments~\cite{clip2video}.
To further examine the effectiveness and potential of CelebV-Text, we evaluate it on a representative baseline \cite{mmvid} for facial text-to-video generation. Our results show better relevance between generated face videos and texts when compared to a state-of-the-art large-scale pretrained model~\cite{cogvideo}. 
Furthermore, we show that a simple modification of \cite{mmvid} with text interpolation can significantly improve temporal coherence. 
Finally, we present a new benchmark for text-to-video generation to standardize the facial text-to-video generation task, which includes representative models \cite{tfgan, mmvid} on three text-video datasets.

The main contributions of this work are summarized as follows:
1) We propose CelebV-Text, the first large-scale facial text-video dataset with high-quality videos, as well as rich and highly-relevant texts, to facilitate research in facial text-to-video generation.
% The texts include detailed static attributes and densely-annotated dynamic attributes.
2) Comprehensive statistical analyses are conducted to examine video/text quality and diversity, as well as text-video relevance, demonstrating the superiority of CelebV-Text.
3) A series of self-evaluations are performed to demonstrate the effectiveness and potential of CelebV-Text.
4) A new benchmark for text-to-video generation is constructed to promote the standardization of the facial text-to-video generation task.

% 3) We conduct experiments on text-to-video generation task and construct a corresponding benchmark on face domain, which demonstrates the effectiveness and potential of CelebV-Text.
% o show output samples can benefit from our input texts, which demonstrates the effectiveness and potential of CelebV-Text. Moreover, construct a benchmark given recent state-of-the-art methods and 

% \wayne{6. contribution ($<$ 15 lines)}

\begin{table*}[htb]
\vspace{-3mm}
\caption{\textbf{In-the-wild face video dataset comparison.}  The symbol ``\#'' indicates the number. The abbreviations ``Res.'', ``Dura.'',  ``App.'', ``Cond.'', ``Act.'', ``Emo.'', and ``Dir.'' stand for Resolution, Duration, Appearance, Condition, Action, Emotion, and Direction, respectively. The ``half checkmark'' denotes that CelebV-HQ consists of action attributes with no timestamp.}
\vspace{-2mm}
    \centering
    \resizebox{0.95\textwidth}{!}{
    \begin{tabular}{l|ccc|ccc|ccc|cc} 
    \toprule
    \multicolumn{1}{l|}{\multirow{3}{*}{Datasets}} & \multicolumn{3}{c|}{\multirow{2}{*}{Meta Information}} & \multicolumn{6}{c|}{Attribute Labels} & \multicolumn{2}{c}{\multirow{2}{*}{Text}} \\ \cline{5-10}
\multicolumn{1}{c|}{} & \multicolumn{3}{c|}{} & \multicolumn{3}{c|}{Static} & \multicolumn{3}{c|}{Dynamic} & \multicolumn{2}{l}{} \\ \cline{2-12}
\multicolumn{1}{c|}{} & \#Samples & Res. & Dura. & \multicolumn{1}{c}{\begin{tabular}[c]{@{}c@{}}General\\App.\end{tabular}} & \begin{tabular}[c]{@{}l@{}}Detail\\App.\end{tabular} & \multicolumn{1}{c|}{\begin{tabular}[c]{@{}c@{}}Light\\Cond.\end{tabular}} & Act. & Emo. & \begin{tabular}[c]{@{}c@{}}Light\\Dir.\end{tabular} & Auto & \begin{tabular}[c]{@{}l@{}}Manual\\\end{tabular} \\ 
\midrule
CelebV~\cite{celebv} & 5 & 256$\times$256 & 2hrs & \multicolumn{3}{c|}{\multirow{2}{*}{\backslashbox{}{}}} & \multicolumn{3}{c|}{\multirow{2}{*}{\backslashbox{}{}}} & \multicolumn{2}{c}{\multirow{2}{*}{\backslashbox{}{}}} \\
% Face Forensics~\cite{faceforensics} & 1,004 & 256\times256 & 4hrs & \multicolumn{3}{l|}{} & \multicolumn{3}{c|}{} &  &  \\
% VoxCeleb~\cite{voxceleb} & 21,245 & 224\times224 & 352hrs & \multicolumn{3}{l|}{} & \multicolumn{3}{c|}{} &  &  \\
VoxCeleb2~\cite{voxceleb2} & 150,480 & 224$\times$224 & 2442hrs & \multicolumn{3}{l|}{} & \multicolumn{3}{c|}{} &  &  \\ 
\midrule
CelebV-HQ~\cite{celebvhq} & 35,666 & 512$\times$ 512 & 68hrs & \color{red}{\cmark} & \xmark & \xmark & \halfcheckmark & \color{red}{\cmark} & \xmark & \xmark & \xmark \\
MM-Vox~\cite{mmvid} & 19,522 & 224$\times$224 & 323hrs & \color{red}{\cmark} & \xmark & \xmark & \xmark & \xmark & \xmark & \color{red}{\cmark} & \xmark \\ \midrule
\textbf{CelebV-Text} & 70,000 & 512$\times$512+ & 279hrs & \color{red}{\cmark} & \color{red}{\cmark} & \color{red}{\cmark} & \color{red}{\cmark} & \color{red}{\cmark} & \color{red}{\cmark} & \color{red}{\cmark} & \color{red}{\cmark} \\
\bottomrule
\end{tabular}
}
\vspace{-2mm}
\label{tab:meta}
\end{table*}

\section{Related Work}

%\subsection{Text-to-Video Generation}
\noindent\textbf{Text-to-Video Generation.}
\if 0
Text-driven video generation, which aims at generating videos from text descriptions, is a challenging task and has gained much interest recently.
Mittal \etal.~\cite{kth} first introduced this task to generate semantically consistent videos conditioned on encoded captions.
Other works \cite{IRC-GAN, tfgan, pan2017create} try to generate video samples conditioned on encoded text inputs.
However, due to the low richness of text description and the small number of data samples, the generated video samples are usually at low resolution or pose weak relevance with input texts.
% CogVideo [27], Phenaki [57], Imagen Video [25],  Make a video
Recently, more works~\cite{godiva, mmvid, phenaki, cogvideo, makeitmove, nuwa, nuwa-inf} use discrete latent codes~\cite{taming, vqvae} for more realistic video generation. 
% For example, MMVID~\cite{mmvid} is capable of generating face videos using multimodal inputs \ie, text, segmentation maps, etc., and CogVideo~\cite{cogvideo} a is a large-scale pretrained text-to-video generation model that takes advantage of the its image version~\cite{cogview2}. 
Some works treat videos as a sequence of independent images~\cite{nuwa, nuwa-inf, cogvideo, mmvid}, while Phenaki~\cite{phenaki} considers temporal relations between each frame for a more robust video decoding process. 
Another branch of works leverage diffusion models for text-to-video generation~\cite{vdm,harvey2022flexible,imagen-video,makeavideo}, which require millions or billions of samples to achieve high-quality generation.
\fi
Text-driven video generation, which involves generating videos from text descriptions, has recently gained significant interest as a challenging task. Mittal \etal.~\cite{kth} first introduced this task to generate semantically consistent videos conditioned on encoded captions. Other studies, such as \cite{IRC-GAN, tfgan, pan2017create}, attempt to generate video samples conditioned on encoded text inputs. However, due to the low richness of text descriptions and the small number of data samples, the generated video samples are often at low resolution or lack relevance with the input texts.
More recently, several works~\cite{godiva, mmvid, phenaki, cogvideo, makeitmove, nuwa, nuwa-inf} have employed discrete latent codes~\cite{taming, vqvae} for more realistic video generation. Some of these works treat videos as a sequence of independent images~\cite{nuwa, nuwa-inf, cogvideo, mmvid}, while Phenaki~\cite{phenaki} considers temporal relations between each frame for a more robust video decoding process. Another branch of studies leverage diffusion models for text-to-video generation~\cite{vdm,harvey2022flexible,imagen-video,makeavideo}, which require millions or billions of samples to achieve high-quality generation.
While text-to-video generation methods are rapidly evolving, they are generally designed for generating general videos. Among these methods, only MMVID~\cite{mmvid} has conducted specific experiments with face-centric descriptions. One possible reason for this is that facial text-to-video generation requires more accurate and detailed text descriptions than general tasks. However, there is currently no suitable dataset available that provides such properties for face-centric text-to-video generation.

\if 0
Although text-to-video generation methods are rapidly evolving, they are designed for general videos. Among these methods, only MMVID~\cite{mmvid} has done specific experiments with face-centric descriptions.
The possible reason is that facial text-to-video generation requires more accurate and detailed text descriptions than general tasks. However, a suitable dataset with such properties is still missing.
% However, these models require millions or billions samples to achieve high-quality generation and the have relative low relevance between text and generated videos as the datasets they used are usually noisy.

% \wayne{one paragraph for general, then one paragraph face-centric}
% \zhuhao{there's no face-centric method. only mmvid conducted the face experiments, but not focused on text to face. }
% 
\fi

%\subsection{Multimodal Datasets}
\noindent\textbf{Multimodal Datasets.}
% add MUGEN 
Existing multimodal datasets can be categorized into two classes: open-world and closed-world.
Open-world datasets~\cite{mscoco,laion400,cc3m,MSVD,dedimo,msr-vtt,activitynet,ucook2,howto100m,epic-kitchen,iper,kth} are widely used for text-to-image/video generation tasks. Some of them have manual annotations~\cite{mscoco,cc3m,msr-vtt,activitynet,epic-kitchen} and part of them are directly collected from the Internet, such as subtitles~\cite{laion400,howto100m}. 
% such as MS-COCO~\cite{mscoco}, LAION-400M~\cite{laion400}, and CC3M ~\cite{cc3m}, and MSR-VTT~\cite{msr-vtt}.
% such as CLEVR~\cite{clevr} and MUGEN~\cite{mugen} 
Closed-world datasets are mostly composed of images or videos collected in constrained environment with corresponding information such as text.
% Close-world dataset such as CLEVR~\cite{clevr} and MUGEN~\cite{mugen} are consisted of synthesized images or video with corresponding text, etc. information. 
CLEVR~\cite{clevr} is a synthetic text-image dataset produced by arranging 3D objects with different shapes under a controlled background. 
% to diagnose visual reasoning abilities of deep learning models. 
While MUGEN~\cite{mugen} is a video-audio-text dataset that was collected using CoinRun~\cite{coinrun} by introducing audio and new interactions. The corresponding text is produced by human annotators and grammar templates. 

Multimodal face datasets also exist.
% VoxCeleb~\cite{voxceleb} and VoxCeleb2~\cite{voxceleb2} are close-world audio-video datasets that were originally released for speaker recognition, and further stimulated the development of audiovisual speaker separation and talking face generation domains.
Modified MUG~\cite{mug} is a closed-world text-video dataset that contains 1,039 videos with subjects showing different emotions, where the text descriptions are generated from facial emotions using a fixed template~\cite{tivgan}.
MM-Vox~\cite{mmvid} contains 19,522 face videos from VoxCeleb~\cite{voxceleb}, with 36 facial attributes manually labeled following CelebA~\cite{celeba} and text descriptions generated via Probabilistic Context-Free Grammar (PCFG)~\cite{tedigan}. 
However, both datasets only contain language descriptions related to static facial attributes without considering the temporal state change (\ie, emotion or action) presented in the original face videos. Moreover, the limited label annotations restrict the diversity of the text descriptions, making them sub-optimal for studying the text-to-video generation task on the face domain.
CelebV-HQ~\cite{celebvhq} is the latest high-quality face video dataset that covers facial annotations, including appearance, movement, and emotion. However, it only provides discrete labels and timestamps, with no text descriptions.
%A facial text-video dataset with fine-grained labels and descriptions of the dynamic state change is urgently needed to be proposed.

\section{Dataset Construction}
\begin{figure}
\centering
\includegraphics[width=\linewidth]{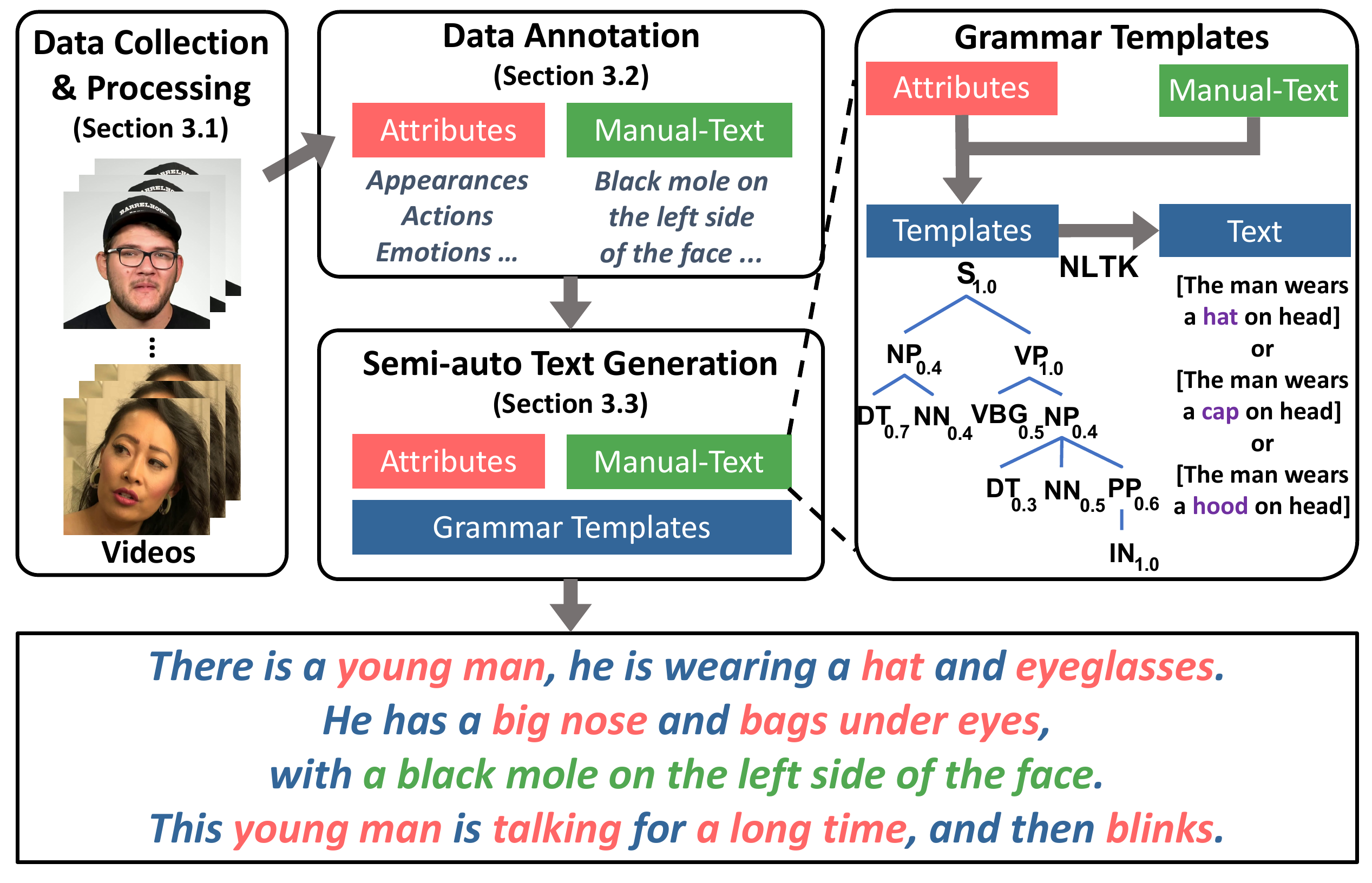}
\caption{\textbf{Pipeline of our dataset construction process.} The pipeline includes data collection \& processing, data annotation, and semi-auto text generation.}
\label{fig:text_pipeline}
\vspace{-5mm}
\end{figure}

In this work, we aim to build a facial text-video dataset, which requires not only large-scale video samples of high quality, but also natural and diverse text descriptions that are highly relevant to videos. 
To achieve so, we propose an efficient pipeline, as shown in Figure~\ref{fig:text_pipeline}, to construct CelebV-Text, including Data Collection \& Processing, Data Annotation, and Semi-auto Text Generation.

% Dataset Construction Pipeline

\subsection{Data Collection \& Processing} \label{sec:3.1}
\noindent{\textbf{Collection.}}
We follow the same strategy as CelebV-HQ~\cite{celebvhq} due to its effectiveness in large-scale high-quality data collection.
Specifically, we firstly generate a large number of queries, including human names, movie titles, vlogs and so on, to retrieve videos that contain human faces with temporally dynamic state changes and abundant facial attributes.
% also filter out videos from MM-vox
% \zhuhao{To to increase the video diversity, we also filter out the videos appearing in CelebV-HQ.}
Our data are collected from open world with videos downloaded from online resources.
Videos with low resolution ($<512^{2}$), low time duration ($<5$s), and having appeared in CelebV-HQ are filtered out.

\noindent{\textbf{Processing.}}
To sample high-quality and diverse video clips from our raw collections, similar steps are followed as CelebV-HQ~\cite{celebvhq} with modifications.
We first filter out video clips with bounding box regions less than $512^{2}$ rather than resize them to the same resolution.
In this way, clips are not upsampled or downsampled hence the video quality would not be affected, which leads to various resolutions of collected videos: $56.4\%$ with $512^{2}\sim1024^{2}$, and $43.6\%$ for $1024^{2}+$.
To reduce the face area noise when the background changes, we further change the video splitting strategy. In addition to our focus on the same human motion~\cite{bewley2016simple} and identity~\cite{arcface} present in adjacent frames, we split the video into different clips when the background changes by a toolkit~\footnote{\url{https://github.com/Breakthrough/PySceneDetect}}.

\subsection{Data Annotation}

The annotation process is a core part in CelebV-Text construction, which would greatly affect the relevance of text-video pairs, as our designed text templates heavily depend on the annotation results. 
Here, we first describe how we design attributes, and then give details about the annotation strategy for face videos.

\begin{figure*}[tb]
    \centering    
    \vspace{-4mm}
    \includegraphics[width=0.90\textwidth]{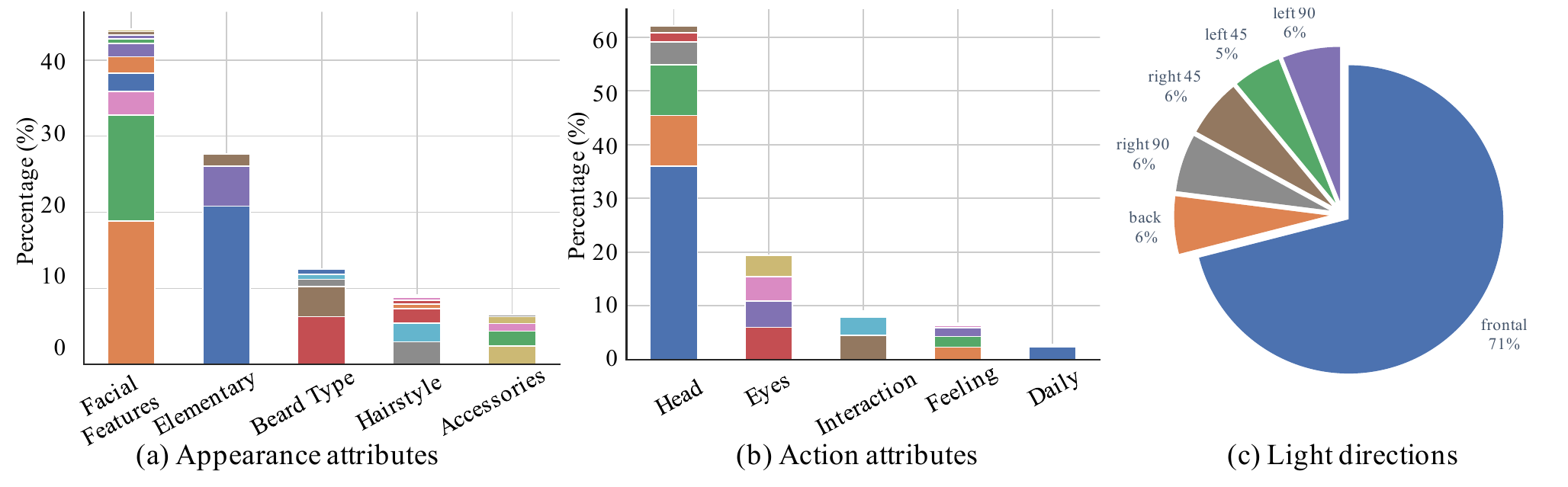}
    \vspace{-4mm}
    \caption{\textbf{Dataset distribution comparison.} The distributions of appearance attributes, action attributes, and light directions. }
    \label{fig:attr_dist}
    \vspace{-5mm}
\end{figure*}

\noindent
\textbf{Attributes Design.}
% \jh{
Temporal dynamic is the key difference between images and videos.
However, as shown in Table~\ref{tab:meta}, most face video datasets focus on static attributes where attribute information does not change over time, such as appearance.
Dynamic attributes that change over time, such as emotion and face actions, are often neglected.
In the following, we decouple face videos into static and dynamic categories and details are given as follows.

\noindent{\textit{1) Static.}}
The current dataset~\cite{mmvid} only considers static information such as the appearance attribute, which includes $40$ classes as CelebA~\cite{celeba}.
In contrast, we define static information to include three types of attributes: general appearance, detailed appearance, and light conditions.
General appearance attributes follow the same definition as CelebA~\cite{celeba}.
Detailed appearance attributes including five classes are proposed for realistic face generation, \ie, scar, mole, freckle, dimple, and one-eyed.
We define light conditions in a restricted manner to include light color temperature~\cite{cct2} and brightness~\cite{brightness}, with a total of $6$ classes.

\noindent{\textit{2) Dynamic.}}
Here, we design three dynamic attributes, \ie, action, emotion, and light directions.
For action attributes, we follow CelebV-HQ~\cite{celebvhq} and expand their action list by two classes, \ie, squint and blink.
For emotion attributes, we select the $8$ emotion setting in Affectnet~\cite{affectnet}, including neutral, anger, contempt, disgust, fear, happiness, sadness, and surprise.
For light direction attributes, we derive and modify classes from~\cite{kan2019deeplight} and give $6$ light direction classes.
Complete lists are given in the Appendix.
Moreover, as shown in Table~\ref{tab:meta}, CelebV-HQ~\cite{celebvhq} is the only dataset giving timestamps of dynamic attributes.
Following their idea, we densely annotate all dynamic attributes of CelebV-Text with the start and end time.

\noindent
\textbf{Automatic and Manual Annotation.}
\label{sec:subsubanno}
Based on our attributes design, we find that some attributes can be annotated automatically (\eg, appearance) while some need manual annotations (\eg, timestamps of dynamic attributes).
Considering the dataset quality and cost of expense, our annotation strategy includes both automatic and manual annotations. 

For automatic annotation, we first investigate algorithms and select designed attributes that can be automatically annotated.
We then test different algorithms on our dataset and keep those giving annotation accuracy of $85\%$ or higher. 
This process yields all light condition labels, all appearance labels, and all emotion labels suitable for automatic annotation.
Algorithms for different labels we finally chose are reported in the Appendix. 
Automatic annotation results can be further revised by human workers to improve accuracy in a less costly way.

For manual annotation, we hire and train human workers following~\cite{celebvhq} to annotate attributes that are filtered out by an automatic annotation process.
In this case, we manually annotate dynamic attributes, \ie, action and light directions, to give both class labels and exact timestamps.
In addition, it is hard to represent detailed appearance attributes by the discrete label, \eg, the characteristics of scars or moles.
We therefore ask annotators to give a natural description for each attribute, describing exact positions relative to face parts.
These designs greatly enhance the relevance between the final text and the video.
% }

\subsection{Semi-auto Text Generation}
Multimodal text-video datasets collect texts via three common methods: subtitles~\cite{frozen, howto100m, crosstask}, manual-text generation~\cite{vatex, activitynet, dedimo, MSVD, msr-vtt}, and auto-text generation~\cite{mugen, tfgan, makeitmove}.
% \jh{
However, it is difficult for the individual method to generate texts with high relevance to videos, natural expression, and high diversity.
Specifically, although subtitles are easy to obtain, they can pose weakly relevant text-video pairs and introduce noise, making the dataset quality hard to control. 
Moreover, manual-text generation method is time and cost consuming, as natural language descriptions are required for each video. 
In this case, increasing the data scale is quite hard as more workers are needed to describe new videos, which does not meet the efficiency and scalability of annotation.
Finally, auto-text generation is flexible and scalable, as abundant texts can be simultaneously generated given annotation results of collected videos.
However, the diversity, complexity, and naturalness of generated texts can be impacted by the designed grammar templates.

To this end, we propose a semi-auto template-based text generation strategy that combines both manual-text and auto-text generation methods. 
% \jh{
Specifically, as mentioned in Section~\ref{sec:subsubanno}, manual-texts are required to describe detailed appearance attributes.
Annotated attribute information is fed into our designed template for auto-text generation.

To make our template as natural as possible, we first ask each annotator to describe 10 different face videos for each attribute.
We then analyze the grammar structure (\ie, parse tree banks) along with online corpora following~\cite{klein2003accurate, nlpparser}, and find the most three common grammar structures for each attribute.
Finally, we utilize probabilistic context-free grammar~\cite{stap2020conditional, tedigan} and modify the grammar structures to design our own templates.
Texts are generated based on templates with synonym replacement using NLTK~\cite{nltk} to increase our generation diversity. Details of our template designs are in the Appendix.

\section{Statistical Analysis of CelebV-Text} 
In this section, we compare CelebV-Text with the two most relevant and representative face video datasets~\cite{mmvid, celebvhq}.
We perform a comprehensive analysis of CelebV-Text in terms of video, text, and text-video relevance.
To verify the effectiveness of our designed grammar templates, we generate text descriptions for CelebV-HQ based on its attributes for comparison.
For simplicity, we use ``CelebV-HQ'' to denote this variant in the following.
% }

\begin{figure}[t]
    \centering
    \vspace{-3mm}
    \includegraphics[width=0.42\textwidth]{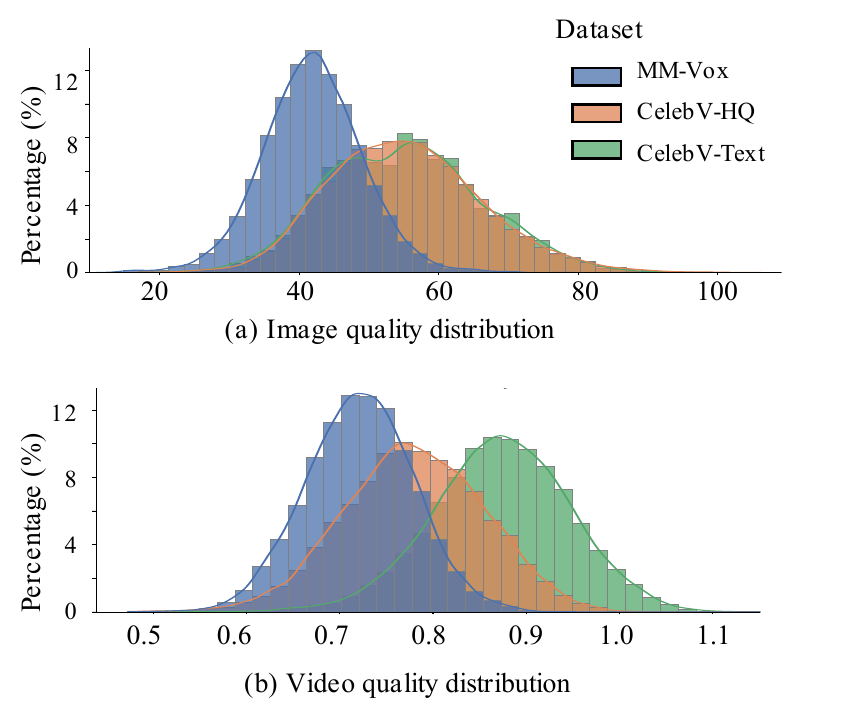}
    \vspace{-5mm}
    \caption{\textbf{Dataset quality distribution.} The metrics used are BRISQUE~\cite{BRISQUE} and VSFA~\cite{VSFA} respectively.}
    \label{fig:quality_dist}
    \vspace{-5mm}
\end{figure}

\subsection{Video Comparisons}
We briefly compare the overall statistics of existing face video datasets~\cite{celebv,voxceleb2,celebvhq,mmvid} in Table~\ref{tab:meta}.
As reported, CelebV-Text contains $70,000$ video clips with a total duration of around $279$ hours.
Each video is accompanied by $20$ sentences describing all $6$ designed attributes.
Compared to CelebV~\cite{celebv}, CelebV-Text has a larger scale and higher resolution.
Although VoxCeleb2~\cite{voxceleb2} has more samples than CelebV-Text, its video distribution is limited as most videos are mainly talking faces.
Moreover, video samples of both CelebV-HQ~\cite{celebvhq} and CelebV-Text are collected in open-world with diverse queries so that they are rich in distribution, while CelebV-Text has about $2$ times video data, more video attributes, and highly relevant text descriptions.
Finally, compared to the only existing facial text-video dataset MM-Vox~\cite{mmvid}, CelebV-Text overpasses MM-Vox in terms of scale and quality.

\begin{table*}[htb!]
\caption{\textbf{Multimodal retrieval results.} Clip2Video~\cite{clip2video} is leveraged to measure the text-video relevance via retrieval experiments. Bold values indicate the best results, underlined ones indicate the second best.}
\vspace{-2mm}
\centering
\resizebox{1\textwidth}{!}{
\begin{tabular}{l|l|lllll|lllll}
\toprule
&  & \multicolumn{5}{c|}{\textbf{Text $\Rightarrow$ Video}} & \multicolumn{5}{c}{\textbf{Video $\Rightarrow$ Text}} \\ \midrule
Description & Dataset & R@1($\uparrow$) & R@5($\uparrow$) & \multicolumn{1}{l|}{R@10($\uparrow$)} & MdR($\downarrow$) & MnR$(\downarrow$) & R@1($\uparrow$) & R@5($\uparrow$) & \multicolumn{1}{l|}{R@10($\uparrow$)} & MdR($\downarrow$) & MnR($\downarrow$) \\ \midrule
\multirow{3}{*}{(a) App.} & MM-Vox~\cite{mmvid} & 1.5 & 9.0 & \multicolumn{1}{l|}{15.7} & 52.0 & 68.8 & 2.0 & 9.2 & \multicolumn{1}{l|}{14.6} & 43.0 & 57.8 \\
 & CelebV-HQ~\cite{celebvhq} & 5.9 & 19.2 & \multicolumn{1}{l|}{29.7} & 27.0 & 52.2 & 7.2 & 20.7 & \multicolumn{1}{l|}{32.4} & 27.0 & 46.9 \\
 & CelebV-Text & 6.1 & 21.3 & \multicolumn{1}{l|}{35.5} & 26.3 & 49.1 & 7.4 & 20.7 & \multicolumn{1}{l|}{29.9} & 26.6 & 48.3 \\ \midrule
\multirow{2}{*}{\begin{tabular}[c]{@{}l@{}}(b) App.+Emo.\end{tabular}} & CelebV-HQ~\cite{celebvhq} & 6.5 & 20.1 & \multicolumn{1}{l|}{30.8} & \textbf{25.0} & 48.0 & 7.9 & 25.5 & \multicolumn{1}{l|}{\textbf{38.8}} & \underline{17.0} & \underline{37.0} \\
 & CelebV-Text & \underline{6.6} & \underline{23.4} & \multicolumn{1}{l|}{\underline{37.1}} & 26.0 & \underline{47.6} & \textbf{8.1} & \underline{27.2} & \multicolumn{1}{l|}{34.7} & 18.2 & 38.3 \\ \midrule
\begin{tabular}[c]{@{}l@{}}(c) App.+Emo.+Act.\end{tabular} & CelebV-Text & \textbf{6.9} & \textbf{24.1} & \multicolumn{1}{l|}{\textbf{39.2}} & \underline{25.8} & \textbf{46.7} & \underline{8.0} &\textbf{27.6} & \multicolumn{1}{l|}{\underline{37.1}} & \textbf{16.7} & \textbf{36.1} \\ \bottomrule
\end{tabular}}
\vspace{-3mm}
\label{tab:ret}
\end{table*}

\begin{figure}[t]
    \centering
    \vspace{-3mm}
    \includegraphics[width=0.95\linewidth]{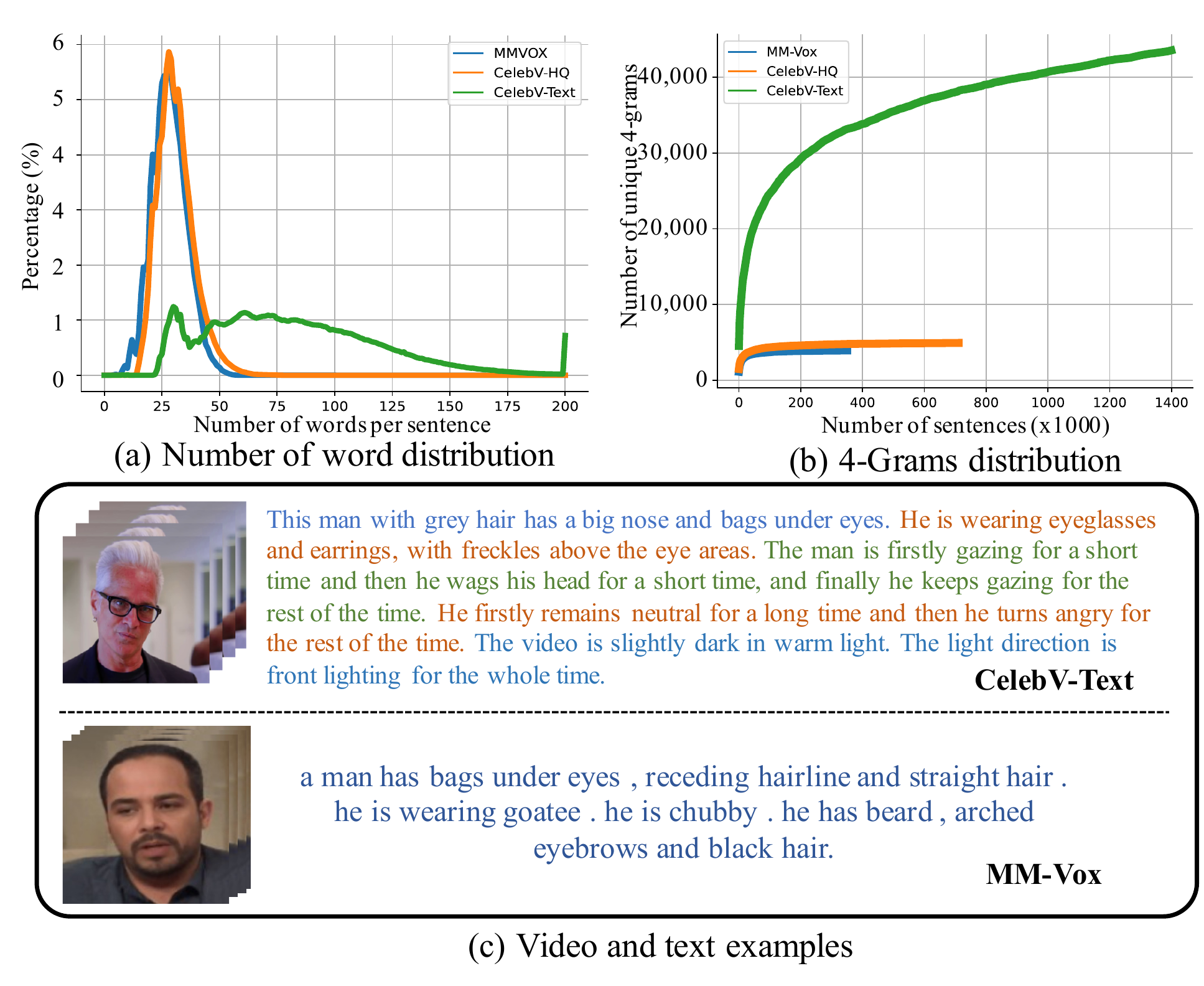}
    \vspace{-4mm}
    \caption{\textbf{Text distribution.} CelebV-Text achieves better performance in both $4$-gram and number words distribution. }
    \label{fig:text_dist}
    \vspace{-2mm}
\end{figure}

\noindent\textbf{Attributes Distribution.}
In order to better present the distribution of different attributes in CelebV-Text, we pick and divide general appearance, action, and light direction attributes into groups.
More distributions and division designs are provided in the Appendix.
Specifically, all $40$ general appearance classes are divided into $5$ groups shown in Figure~\ref{fig:attr_dist}~(a).
Facial features (\eg, double chin, big nose, and oval face) account for the most portion around $45\%$.
The elementary group is twice large than the beard type, accounting for around 25\% and $12\%$, respectively.
Fewer samples are located to the hairstyle and accessories groups, taking around $10\%$ and $8\%$, respectively.
% action
Besides, action attributes are divided into $5$ groups in Figure~\ref{fig:attr_dist}~(b), where it is clear that head-related actions account for the largest portion of around $60\%$, followed by eyes-related actions of around $20\%$.
The interaction group (\eg, eat), feeling group (\eg, smile), and daily group (\eg, sleep) account for around $9\%$, $7\%$, and $4\%$, respectively.
%light
Finally, for light directions (Figure~\ref{fig:attr_dist}~(c)), most samples contain the front lighting and the remaining ones are evenly distributed.
% } 

\begin{table}[t]
\footnotesize
\caption{\textbf{Number of unique POS tags.} The numbers of unique POS tags for MM-Vox, CelebV-HQ, and CelebV-Text.}
\vspace{-2mm}
\centering
\resizebox{0.40\textwidth}{!}{
\begin{tabular}{l|cccc}
\toprule
%  & \multicolumn{4}{c}{\#unique \textit{n}-grams} \\ \hline
% \multicolumn{1}{l|}{Dataset} & \multicolumn{1}{c}{1-grams} & \multicolumn{1}{c}{2-grams} & \multicolumn{1}{c}{3-grams} & \multicolumn{1}{c}{4-grams} \\ \midrule
% MM-Vox~\cite{mmvid} & 65 & 243 & 1478 & 3935 \\
% CelebV-HQ~\cite{celebvhq} & 103 & 372 & 1866 & 4932 \\
% CelebV-Text & \textbf{593} & \textbf{3385} & \textbf{14136} & \textbf{45,692} \\ \midrule
 % & \multicolumn{4}{c}{\#unique POS Tag} \\ \midrule
Dataset & \multicolumn{1}{c}{\#Verb} & \multicolumn{1}{c}{\#Adj.} & \multicolumn{1}{c}{\#Noun} & \multicolumn{1}{c}{\#Adv.} \\ \midrule
MM-Vox~\cite{mmvid} & 5 & 20 & 38 & 0 \\
CelebV-HQ~\cite{celebvhq} & 10 & 24 & 50 & 6 \\
\textbf{CelebV-Text} & \textbf{96} & \textbf{78} & \textbf{174} & \textbf{24} \\ \bottomrule
\end{tabular}}
% \caption{\textbf{Text distribution.} The numbers of unique n-grams and POS tags for MM-Vox, CelebV-HQ, and CelebV-Text.}
\vspace{-6mm}
\label{tbl:text_dist}
\end{table}

\noindent\textbf{Video Quality Distribution.}
% \zhuhao{Analysis of Figure4}
% \zhuhao{Since CelebV-HQ is the largest high-quality face video dataset available, and we follow the same video acquisition strategy as it. Therefore, we compare the video quality of CelebV-Text with MM-Vox~\misscite and CelebV-HQ~\misscite  to demonstrate the high quality of CelebV-Text. 
% We use the metrics presented in the CelebV-HQ, \ie, Mean BRISQUE~\cite{BRISQUE} for static quality evaluation, and VSFA~\cite{VSFA} for video quality distributions. 
% As shown in Figure~\ref{fig:quality_dist}(a), the static image quality, CelebV-Text and CelebV-HQ achieve comparable quality and are both higher than MM-Vox~\misscite. 
% For video quality, the ranking is CelebV-Text, CelebV-HQ, and MM-Vox, as reported in Figure~\ref{fig:quality_dist}(b). We believe that CelebV-Text is better than CelebV-HQ due to the effect of the video split method we mentioned in Section~\missref, which alleviates the discontinuity caused by background transitions.}
% \jh{
We follow \cite{celebvhq} to analyze the quality of our collected videos.
To demonstrate the superiority of CelebV-Text, we compare with MM-Vox~\cite{mmvid} and CelebV-HQ~\cite{celebvhq}, where mean BRISQUE~\cite{BRISQUE} and VSFA~\cite{VSFA} are used to evaluate the image and video quality, respectively.
Image quality of all datasets is shown in Figure~\ref{fig:quality_dist}~(a), where CelebV-Text and CelebV-HQ achieve comparable quality, higher than MM-Vox by a large margin.
Video quality of all datasets is shown in Figure~\ref{fig:quality_dist}~(b), where CelebV-Text has the best quality, which is due to the effect of the video split method mentioned in Section~\ref{sec:3.1}, alleviating the discontinuity during background transitions.
% }

\subsection{Text Comparisons}
In addition to a large number of video samples, text descriptions of CelebV-Text are longer and more detailed than those in MM-Vox~\cite{mmvid} and CelebV-HQ~\cite{celebvhq} (see Figure~\ref{fig:text_dist}~(a)), where the average text length of MM-Vox, CelebV-HQ, and CelebV-Text are $28.39$, $31.06$, and $67.15$.
Distributions of Celeb-HQ and MM-Vox are close, but there are more words in CelebV-Text to describe a video due to the comprehensive annotation.

To validate the linguistic diversity of the generated texts, comparisons are conducted among the three datasets following~\cite{vatex}.
Specifically, we report the unique part-of-speech (POS) tags (\ie, verb, noun, adjective, and adverb) of the three datasets in Table~\ref{tbl:text_dist}.
Obviously, due to our comprehensively designed attribute list and the number of templates, CelebV-Text presents a wider variety of text styles, covering a broader range of face attributes that are static or dynamic in the temporal domain.

In addition, we further examine the naturalness and complexity of our texts compared to MM-Vox, where we modify~\cite{typetoken} to calculate the type-token vocabulary curve for all captions.
As shown in Figure~\ref{fig:text_dist}~(b) where unique $4$-grams are selected as the types~\cite{vatex}, it is evident that due to our grammar structures and synonym replacement, the linguistic naturalness (vocabulary use) and complexity (vocabulary size) of our CelebV-Text are much better. Please refer to Appendix for more $n$-grams results.

\begin{figure*}
\centering
\vspace{-3mm}
\includegraphics[width=0.95\linewidth]{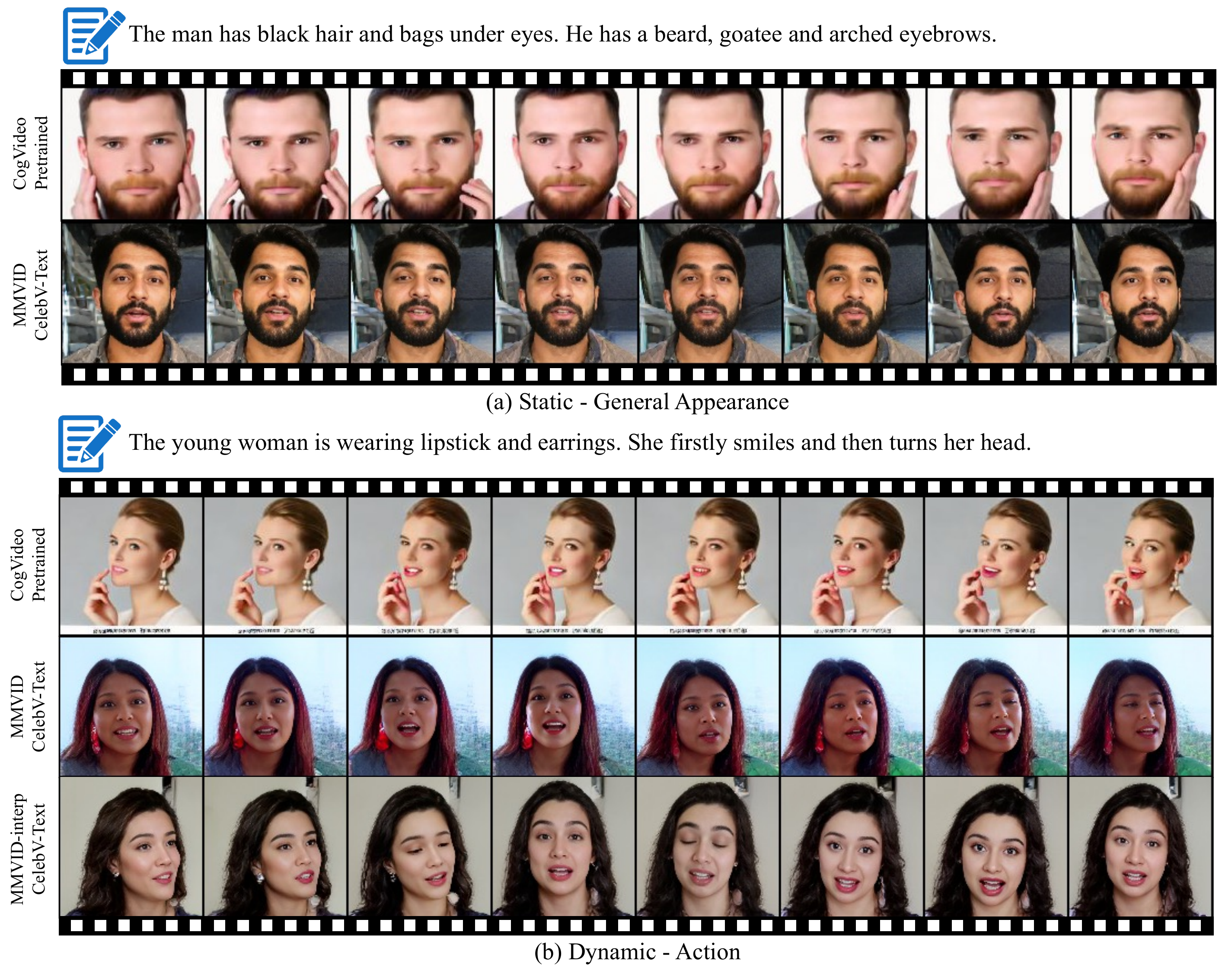}
\vspace{-5mm}
\caption{\textbf{Qualitative results of facial text-to-video generation.} The generated samples are given texts describing (a) the static attribute and (b) dynamic attribute. 
}
\vspace{-2mm}
\label{fig:static_results}
\end{figure*}

\subsection{Text-Video Relevance}
To quantitatively validate our text-video relevance, we conduct text-video retrieval tasks on three datasets: MM-Vox~\cite{mmvid}, CelebV-HQ~\cite{celebvhq}, and CelebV-Text.
Rather than use conventional frame-wise clip score as most works~\cite{phenaki,imagen-video,makeavideo}, we follow~\cite{clip2video} to compute feature similarities between texts and videos with the consideration of temporal dynamics, which reflects accurate multimodal interactions across the two modalities.
Recall at rank K (R@K), median rank (MdR), and mean rank (MnR)~\cite{zhang2018cross,clip2video,mithun2018learning} are used as evaluation metrics, where the higher R@K, the lower median rank and mean rank indicate better performance.

We first examine the performance given texts with descriptions of general appearance in Table~\ref{tab:ret}~(a).
Results of CelebV-HQ and CelebV-Text are both better than MM-Vox for two retrieval tasks, which indicates our designed templates can produce texts more relevant to videos than MM-Vox.
We further add descriptions about dynamic emotion changes to CelebV-HQ and CelebV-Text in Table~\ref{tab:ret}~(b).
Similar results are achieved in both datasets, which reflects that our annotation accuracy on static appearance attributes is as good as CelebV-HQ.
Finally, we append action descriptions to CelebV-Text in Table~\ref{tab:ret}~(c), which achieves the best performance on most metrics, verifying the relevance between our generated texts and video samples.

% quantitative evaluation table
\begin{table*}
\caption{\textbf{Benchmark of text-to-video generation on different datasets.} $\downarrow$ means a lower value is better and $\uparrow$ means the opposite.
}
\vspace{-3mm}

\footnotesize
\centering
\begin{subtable}{0.47\linewidth}
\caption{Quantitative results on general appearance descriptions.} \label{tab:ours}
\resizebox{1\textwidth}{!}{
\begin{tabular}{l|l|l|l|l}
\toprule
Dataset & Method & FVD($\downarrow$) & FID($\downarrow$) & CLIPSIM($\uparrow$) \\ \hline
\multirow{2}{*}{MM-Vox~\cite{mmvid}} & TFGAN~\cite{tfgan} & 502.28 $\pm$ 1.66 & 760.24 $\pm$ 16.01 & 0.165 $\pm$ 0.022 \\ 
 & MMVID~\cite{mmvid} & \textbf{65.79 $\pm$ 1.81} & \textbf{38.81 $\pm$ 3.66} & \textbf{0.170 $\pm$ 0.020} \\ \midrule
\multirow{2}{*}{CelebV-HQ~\cite{celebvhq}} & TFGAN~\cite{tfgan} & 428.04 $\pm$ 1.76 & 616.24 $\pm$ 17.45 & 0.168 $\pm$ 0.021 \\ 
 & MMVID~\cite{mmvid} & \textbf{73.65 $\pm$ 1.43} & \textbf{63.86 $\pm$ 3.66} & \textbf{0.172 $\pm$ 0.019} \\ \midrule
\multirow{2}{*}{\textbf{CelebV-Text}} & TFGAN~\cite{tfgan} & 403.04 $\pm$ 1.34 & 589.24 $\pm$ 16.46 & 0.177 $\pm$ 0.012 \\ 
 & MMVID~\cite{mmvid} & \textbf{66.69 $\pm$ 1.35} & \textbf{58.70 $\pm$ 4.67} & \textbf{0.198 $\pm$ 0.014} \\ \bottomrule
\end{tabular}}
\end{subtable}
     \hfill
\begin{subtable}{0.47\linewidth}
\caption{Quantitative results on dynamic descriptions of CelebV-Text.}
\centering
\resizebox{1\textwidth}{!}{
\begin{tabular}{l|l|l|l|l}
\toprule
Dataset & Method & FVD($\downarrow$) & FID($\downarrow$) & CLIPSIM($\uparrow$) \\ \toprule
\multirow{3}{*}{\begin{tabular}[c]{@{}l@{}} CelebV-Text\\ \textbf{App.+Emo.}\end{tabular}} & TFGAN~\cite{tfgan} & 442.30 $\pm$ 2.56 & 623.17 $\pm$ 18.88 & 0.158 $\pm$ 0.024 \\
 & MMVID~\cite{mmvid} & 82.78 $\pm$ 1.47 & 61.58 $\pm$ 3.99 & 0.176 $\pm$ 0.008 \\
 & MMVID-interp & \textbf{72.87 $\pm$ 1.23} & \textbf{41.57 $\pm$ 3.56} & \textbf{0.182 $\pm$ 0.010} \\ \midrule
\multirow{3}{*}{\begin{tabular}[c]{@{}l@{}}CelebV-Text\\ \textbf{App.+Act.} \end{tabular}} & TFGAN~\cite{tfgan} & 571.34 $\pm$ 4.54 & 784.93 $\pm$ 20.13 & 0.154 $\pm$ 0.028 \\
 & MMVID~\cite{mmvid} & 109.25 $\pm$ 2.11 & 82.55 $\pm$ 4.37 & 0.174 $\pm$ 0.019 \\ 
 & MMVID-interp & \textbf{80.81 $\pm$ 2.55} & \textbf{70.88 $\pm$ 4.77} & \textbf{0.176 $\pm$ 0.020}  \\ \bottomrule
\end{tabular}
}
\end{subtable}
\vspace{-4mm}
\label{tab:benchmark}
\end{table*}

\section{Experiment}
In this section, we first conduct facial text-to-video generation to validate the effectiveness of our CelebV-Text dataset. We then benchmark representative approaches on facial text-to-video generation task.

\subsection{High-relevance Text-to-Video Generation} \label{sec:5.1}
To show the benefits brought by our text descriptions which depict both static and dynamic attributes, we conduct experiments to show the effectiveness of CelebV-Text. Experiments are mainly based on a recent open-sourced state-of-the-art method, MMVID~\cite{mmvid}, and compared with CogVideo\footnote{We choose CogVideo~\cite{cogvideo} as the representative large-scale model for comparison, since the inference code and pretrained models of other large-scale methods (\eg, CogVideo~\cite{cogvideo}, Phenaki~\cite{phenaki}, Imagen Video~\cite{imagen-video}, and Make-A-Video\cite{makeavideo}) are not public.}~\cite{cogvideo}, which is a large-scale pretrained text-to-video model, trained on millions of text-image/video pairs. 

\begin{figure*}[t]

\vspace{-3mm}
    \includegraphics[width=.95\linewidth]{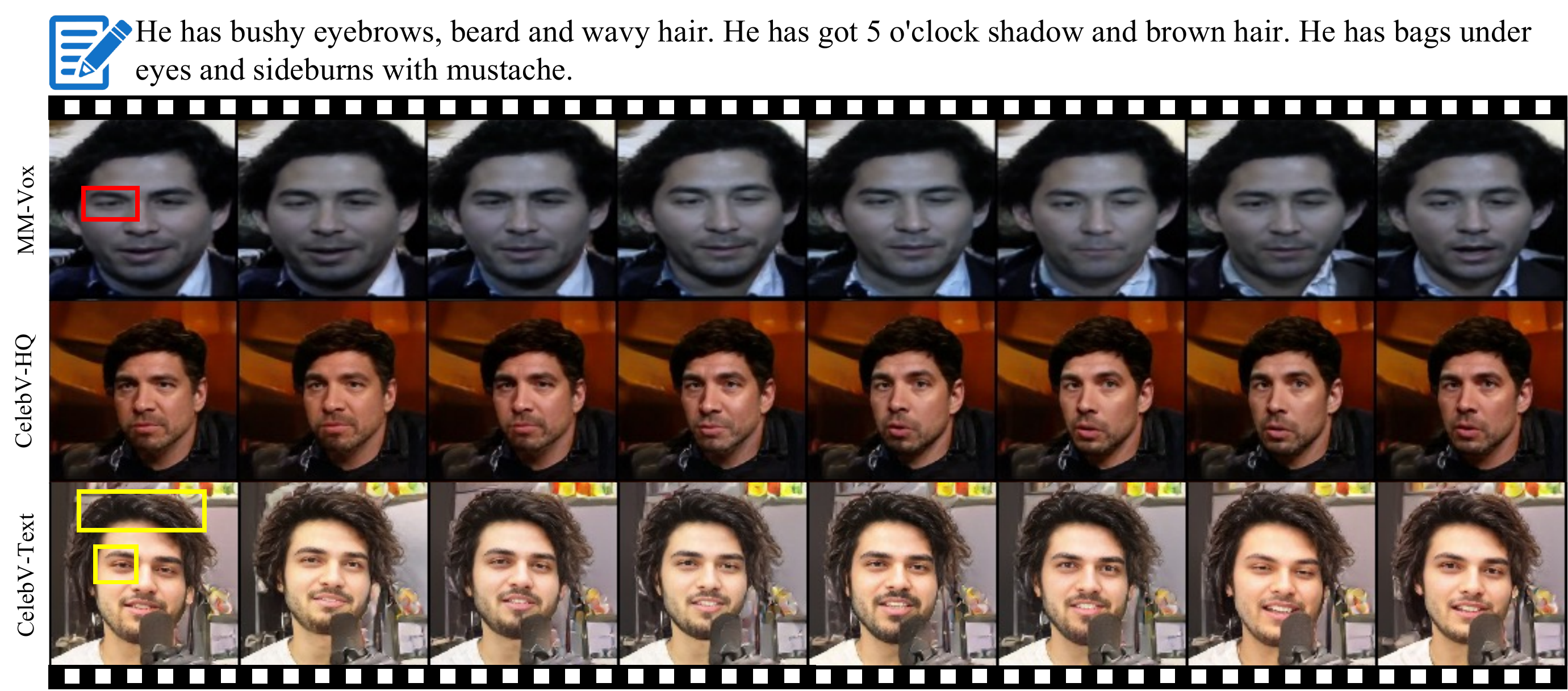}
\vspace{-3mm}
    \caption{\textbf{Qualitative results on three facial text-video datasets.} Red and yellow regions indicate the missing of ``bags under eyes" and the existence of ``wavy hair" and ``bags under eyes".}
    \label{fig:benchmark}
\vspace{-4mm}
\end{figure*}

\noindent\textbf{Static Face Video Generation.}
To validate the effectiveness of our facial text-video dataset in static attributes, we use the models stated above to generate videos conditioned on general appearance, face details, and light conditions descriptions, respectively. 
Specifically, we first train MMVID~\cite{mmvid} from scratch solely on CelebV-Text.
We then generate $3$ input texts including individual descriptions of each of the static attributes.
Generated texts are fed into both MMVID~\cite{mmvid} and CogVideo~\cite{cogvideo} and corresponding video outputs are examined.

Visualization results of general appearance are shown in Figure~\ref{fig:static_results}~(a), which prove the effectiveness of our dataset.
We observe that although CogVideo can output the face video given a text description, the text-video pair is not quite relevant, such as ``bags under eyes'' and  ``wavy hair''.
However, MMVID~\cite{mmvid} produces videos with high relevance to input texts, containing all attributes described in the text.
More results are shown in the Appendix.

% new version of Dynamic Face Video Generation
\noindent\textbf{Dynamic Face Video Generation.}
We follow the above experimental setting to validate the effectiveness of our dataset with dynamic attribute changes (\ie, emotion, action and light direction).
Due to the difficulty in modelling state change~\cite{makeavideo, makeitmove}, we follow \cite{tfgan} to apply test-time interpolation to MMVID~\cite{mmvid}, named MMVID-interp, to improve the text encoding and better understand the dynamics.
Details of our modification are shown in the Appendix.
% We also evaluate the performance of CogVideo and compare MMVID trained on CelebV-Tex` with CogVideo pretrained on a large-scale general datasets in Figure~\ref{fig:temporal_results}.

In Figure~\ref{fig:static_results}~(b), we observe that CogVideo fails to reflect the temporal change described in the input text, \ie, smile $\rightarrow$ turn.
However, both MMVID~\cite{mmvid} and MMVID-interp trained on CelebV-Text can successfully model the dynamic attribute changes, which demonstrates the effectiveness of our dataset.
In addition, we find that MMVID~\cite{mmvid} cannot preserve some attributes well (\eg, earrings), while MMVID-interp can stabilize the sampling process, validating the effectiveness of our modification.
More results are shown in the Appendix.

Note that CogVideo~\cite{cogvideo} has a much larger model size ($\sim100$ times larger than MMVID~\cite{mmvid}) and is trained on much large text-video data ($\sim75$ times larger than CelebV-Text).
However, video samples produced by CogVideo~\cite{cogvideo} shown in Figure~\ref{fig:static_results} are of a lower quality than the ones by MMVID~\cite{mmvid} trained solely on CelebV-Text, where generated faces are not in a high relevance to input texts, demonstrating the effectiveness of our facial text-video dataset.

\subsection{Benchmark on Facial Text-to-Video Generation}
As the domain of text-to-video generation is currently thriving, there exists only one benchmark in the face domain, MM-Vox~\cite{mmvid}.
We expand \cite{mmvid} and construct a benchmark of facial text-to-video generation tasks on three datasets: MM-Vox~\cite{mmvid}, CelebV-HQ~\cite{celebvhq} with texts generated by our templates, and CelebV-Text.
We choose two representative methods\footnote{Other methods, \eg, CogVideo~\cite{cogvideo}, Phenaki~\cite{phenaki}, Imagen Video~\cite{imagen-video}, and Make-A-Video\cite{makeavideo} are not included since their training codes are not public so far.}, TFGAN~\cite{tfgan} and MMVID~\cite{mmvid}, to evaluate their performances on all datasets.

\noindent\textbf{Quantitative Results.}
For thorough benchmark construction, we evaluate baseline methods given variant texts including static and dynamic attributes.
We use FVD~\cite{fvd} (temporal consistency), FID~\cite{fid} (individual frame quality), and CLIPSIM~\cite{godiva} (text-video relevance) as evaluation metrics following \cite{mmvid} and report detailed results for appearance, action, and emotion in Table~\ref{tab:benchmark}.
Evaluation steps are repeated over ten runs with mean values and standard errors reported as well.
All other values are shown in the Appendix.
It can be seen from Table~\ref{tab:benchmark} that MMVID~\cite{mmvid} obtains good FVD/FID/CLIPSIM metrics over TFGAN~\cite{tfgan} which fails to generate reasonable video outputs.
In addition, when input texts contain descriptions about a dynamic state change in the temporal domain, the generated video quality by MMVID~\cite{mmvid} decreases, which encourages future methods to focus more on cross-modal understanding and consistent video generation.
Moreover, the performance of MMVID-interp is better than MMVID~\cite{mmvid} on all metrics, validating the effectiveness of our modification mentioned in Section~\ref{sec:5.1}.
Due to challenges posed by our dataset and text-to-video generation task, there is still considerable room to improve.

\noindent\textbf{Qualitative Results.}
Video samples generated from MMVID~\cite{mmvid} trained on different datasets are shown in Figure~\ref{fig:benchmark}, where all video frames are of $128^{2}$.
We can see that video samples generated by MMVID~\cite{mmvid} trained on different datasets are of high quality with temporal consistency.
However, MMVID~\cite{mmvid} trained on MM-Vox~\cite{mmvid} can sometimes fail to generate attributes mentioned in the input texts.
More generated video samples with dynamic attribute changes are shown in the Appendix.

\section{Discussion}
\if 0
In this work, we propose CelebV-Text, a large-scale, high-quality, and diverse facial text-video dataset with static and dynamic attributes.
CelebV-Text contains $70,000$ video clips, each of which is accompanied by $20$ individual sentences, describing both static and dynamic factors.
Through extensive statistical analysis and experiments, we demonstrate the superiority and effectiveness of CelebV-Text.
In the future, we plan to further enlarge CelebV-Text in both the scale and diversity, to give continuous support to the facial text-to-video generation field. Also, based on CelebV-Text, several new tasks are attractive to explore, \ie, the fine-grained control of video face, adaptation of general pretrained models to the face domain, and text-driven 3D-aware facial video generation.
\fi

We have proposed CelebV-Text, a large-scale, high-quality, and diverse facial text-video dataset with static and dynamic attributes. CelebV-Text contains $70,000$ video clips, each of which is accompanied by $20$ individual sentences describing both static and dynamic factors. Through extensive statistical analysis and experiments, we have demonstrated the superiority and effectiveness of CelebV-Text. In the future, we plan to further enlarge CelebV-Text in both scale and diversity. We may further explore several new tasks based on CelebV-Text, such as fine-grained control of video face, adaptation of general pretrained models to the face domain, and text-driven 3D-aware facial video generation.

% \section*{Ethical Consideration}
\noindent{\textbf{Ethical Consideration.}}
% \textit{(1) License.}
\if 0
CelebV-Text can only be used for research purposes. The raw videos will not be released, while the data annotations, links of raw videos, and data processing tools will be released, following a strict legality check procedure of our institution.
% \textit{(2) Identity information.}
Note that, our data annotation does not include any personal biometric information (\eg, identity), only generic attribute information such as gender, hair color, and motion is annotated.
Moreover, synthetic videos generated in the paper do not show bias or certain biometric information (\eg, big lips or big nose), which alleviates the ethical issues.
% \textit{(3) Misuse.}
CelebV-Text can be used for deepfakes, while it also can be used for the forgery detection task to prevent this issue.
We will strictly control the application and acquisition procedure of CelebV-Text, to avoid possible misuse and abuse.
In the future, we will utilize synthetic face generation framework to generate synthetic face videos to overcome the ethical shortcomings of existing real-world face video datasets.
\fi
%
CelebV-Text is intended for research purposes only. While the raw videos will not be released, the data annotations, links to raw videos, and data processing tools will be made available after undergoing a rigorous legality check procedure at our institution. It is worth noting that our data annotation does not include any personal biometric information such as identity. Only generic attribute information such as gender, hair color, and motion is annotated. Additionally, synthetic videos generated in this work do not exhibit bias or certain biometric information (\eg, big lips or big nose), alleviating ethical concerns. CelebV-Text may be used for deepfakes, but it can also be used for forgery detection tasks to prevent such issues. We will try our best to control the application and acquisition procedure of CelebV-Text to avoid potential misuse and abuse. In the future, we plan to use synthetic face generation frameworks to generate synthetic face videos to address the ethical shortcomings of existing real-world face video datasets.

% \textit{(4) Issues of minions.} 
%\section*{Acknowledgement}
\noindent\textbf{Acknowledgement.}
CelebV-Text is developed under OpenXDLab -- an open platform for X-Dimension high-quality data. This study is supported by the RIE2020 Industry Alignment Fund Industry Collaboration Projects (IAF-ICP) Funding Initiative, in-kind contribution from the industry partner(s), and the MOE AcRF Tier 1 (RG16/21).

%%%%%%%%% REFERENCES
{\small
\bibliographystyle{ieee_fullname}
\bibliography{egbib}
}

\newpage

\appendix

\noindent
\textbf{\LARGE Appendix}

\setcounter{table}{0}
\renewcommand{\thetable}{A\arabic{table}}
\setcounter{figure}{0}
\renewcommand{\thefigure}{A\arabic{figure}}

\section{Details of Attribute Designs}
\subsection{Complete Attribute Lists}
The complete list of all the attributes is reported in Table~\ref{tab:attr_list}.

\subsection{Grouped Attribute Details}
In the main paper, in order to better present the distributions, we divide 40 appearance attributes into facial features, elementary, beard type, hairstyle, and accessories. \\
\textbf{a. Facial features}: double chin, pale skin, high cheekbones, chubby, oval face, bushy eyebrows, bags under eyes, narrow eyes, heavy makeup, arched eyebrows, pointy nose, big nose, big lips.  \\
\textbf{b. Elementary}: young, male, blurry. \\
\textbf{c. Beard type}: 5 o'clock shadow, no beard, goatee, sideburns, mustache. \\
\textbf{d. Hairstyle}: blond hair, gray hair, brown hair, black hair, wavy hair, receding hairline, bangs, straight hair, bald. \\
\textbf{e. Accessories}: wearing earrings, wearing hat, wearing necktie, wearing necklace, eyeglasses, wearing lipstick

Moreover, all 37 actions are split into Head, Eyes, Interaction, Feeling, and Daily groups.  \\
\textbf{a. Head}: talk, head wagging, look around, turn, shake head, nod. \\
\textbf{b. Eyes}: blink, wink, squint, close eyes \\
\textbf{c. Interaction}: drink, sing, eat, smoke, listen to music, play instrument, read, kiss , whisper.  \\
\textbf{d. Feeling}: sneer, sigh, frown, weep, cry, smile, glare, gaze, laugh, shout. \\
\textbf{e. Daily}: yawn, sneeze, cough, sleep, make a face, smoke, blow, sniff, chew.

\subsection{More Distributions}
\begin{figure}
    \centering   
    \includegraphics[width=0.95\linewidth]{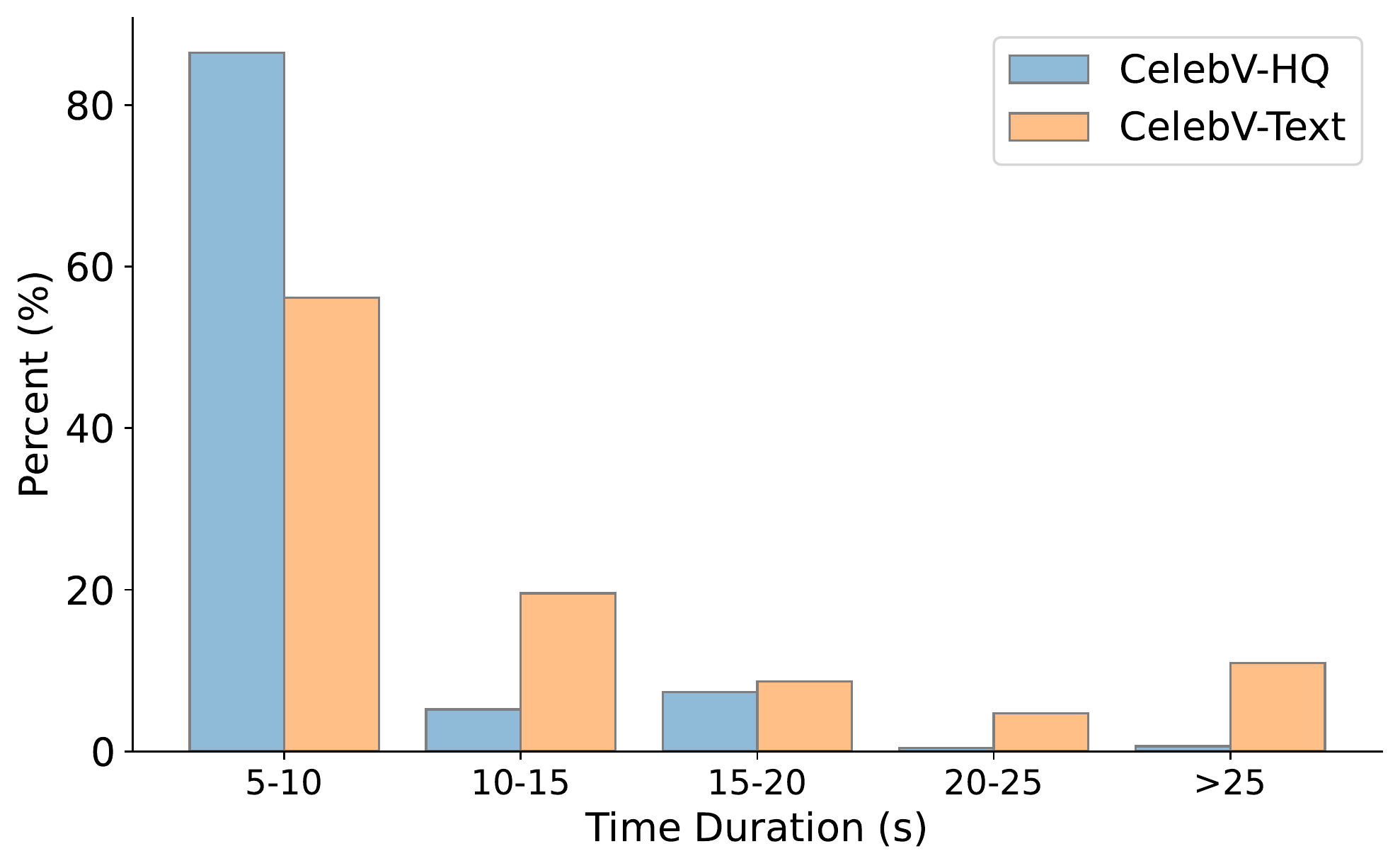}
    \caption{Video time duration of CelebV-Text compared with CelebV-HQ~\cite{celebvhq}.}
    \label{fig:duration}
\end{figure}

\begin{figure*}
    \centering   
    \includegraphics[width=.94\textwidth]{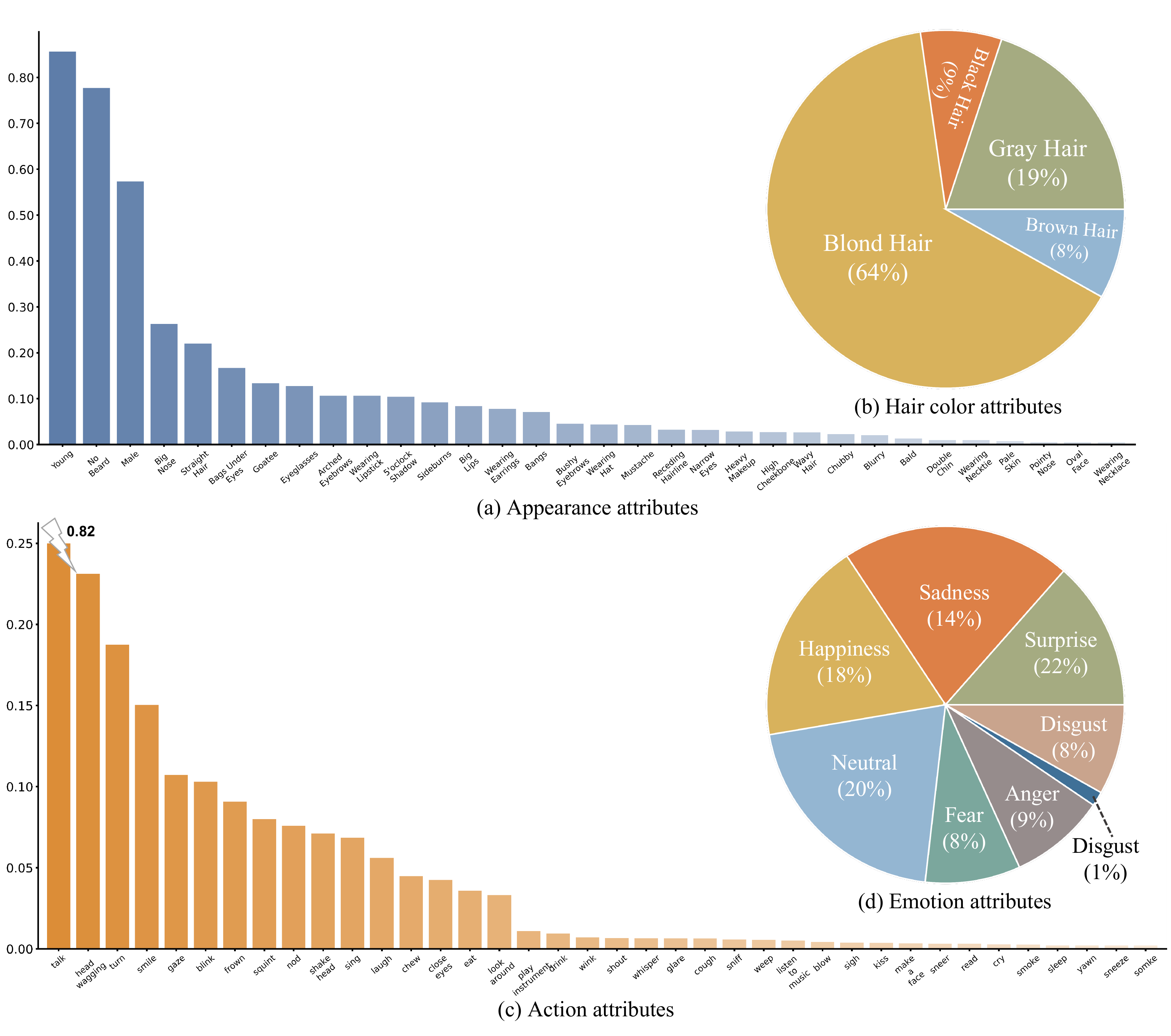}
    \caption{Distributions of general appearances, hair colors, actions, and emotions.}
    \label{fig:general_dist}
\end{figure*}

\begin{figure*}
    \centering   
    \includegraphics[width=0.9\textwidth]{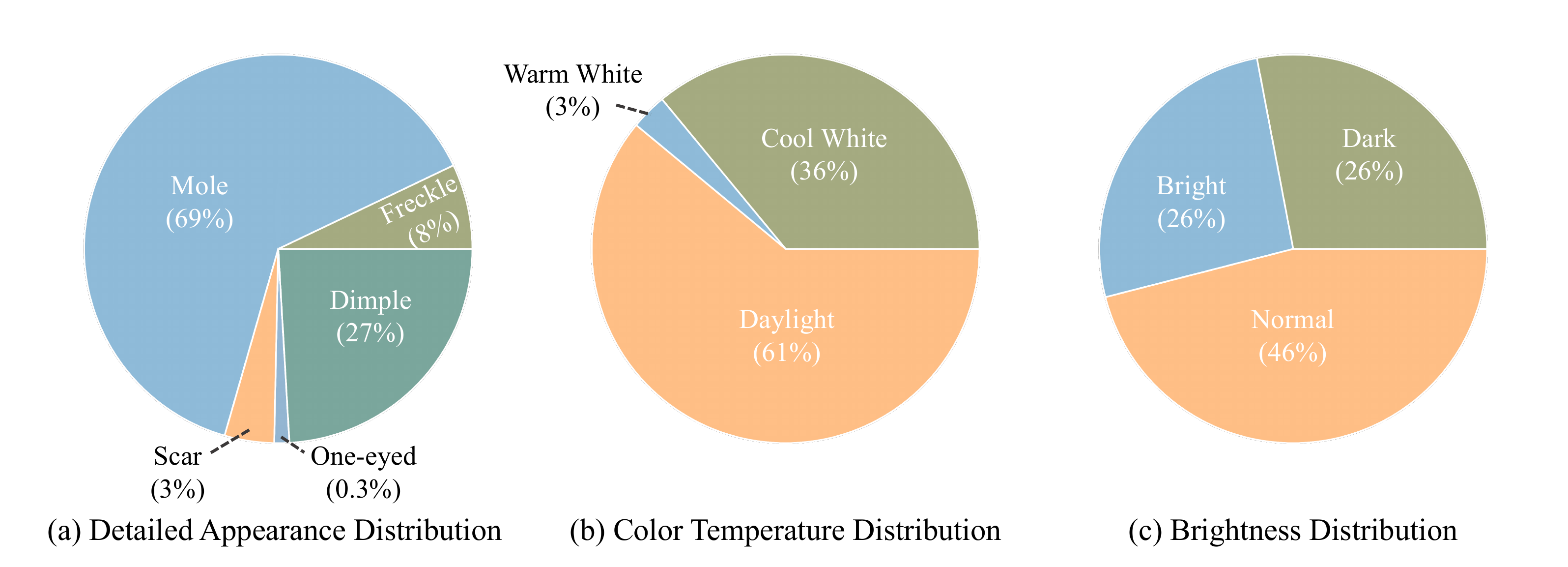}
    \caption{Distributions of detailed appearances, color temperature, and brightness.}
    \label{fig:more_dist}
\end{figure*}

\begin{figure*}
    \centering   
    \includegraphics[width=0.9\textwidth]{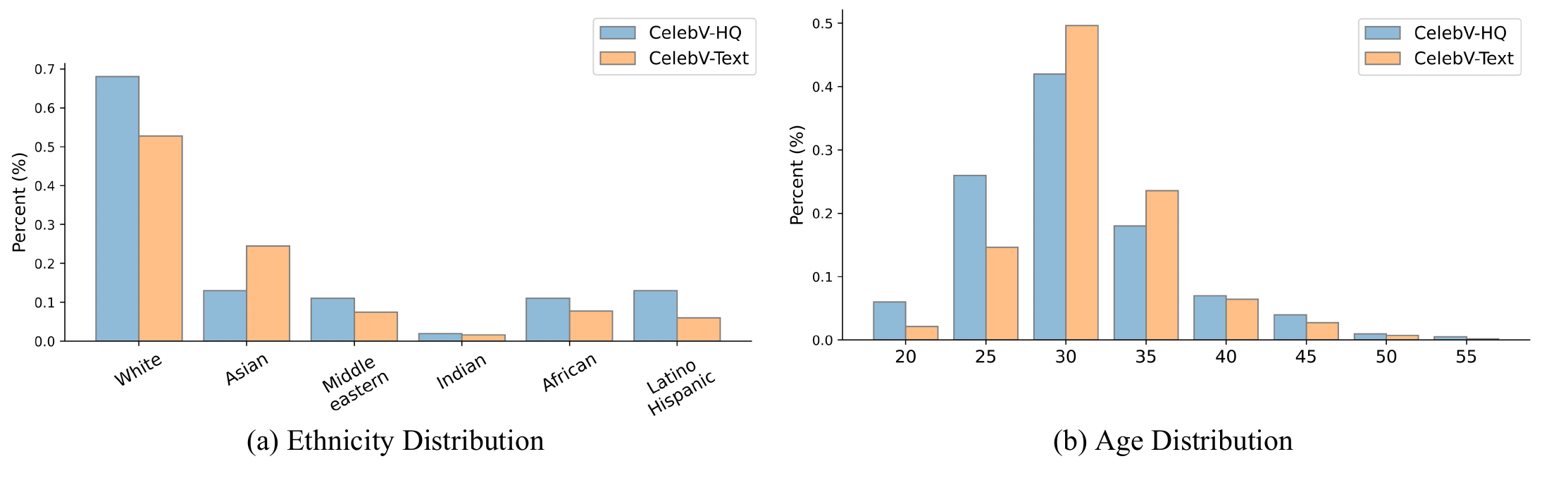}
    \caption{Distributions of ethnicity and age compared with CelebV-HQ~\cite{celebvhq}.}
    \label{fig:race_age}
\end{figure*}

To show the reasonable distribution of CelebV-Text, we first compare the video length duration of our collected videos with CelebV-HQ~\cite{celebvhq} in Figure~\ref{fig:duration}, where video duration in CelebV-Text is longer than CelebV-HQ. Moreover, the average time duration of CelebV-Text is 14.34s, which is twice more than that of CelebV-HQ of 6.68s.
We then present the detailed distributions of general appearances, hair colors, actions and emotions following CelebV-HQ~\cite{celebvhq} in Figure~\ref{fig:general_dist}.
More distributions of detailed appearances, color temperatures, and brightness are shown in Figure~\ref{fig:more_dist}.
Finally, we compare with CelebV-HQ~\cite{celebvhq} in more general attributes such as age and ethnicity. 
Since age and ethnicity labels are not manually annotated, we estimate these two attributes using an off-the-shelf facial attribute analysis framework\footnote{\url{https://github.com/serengil/deepface}}.
As illustrated in Figure~\ref{fig:race_age}, CelebV-Text achieves the distributions close to those of CelebV-HQ.

\begin{table*}
\centering
\caption{\textbf{Complete attribute list.} CelebV-Text contains both static and dynamic attributes, including 40 general appearances, 5 detailed appearances, 6 light conditions, 37 actions, 8 emotions, and 6 light directions.}
\label{tab:attr_list}
\resizebox{0.72\textwidth}{!}{%
\begin{tabular}{cccccc} 
\toprule
\multicolumn{6}{c}{\textbf{Static Attributes}}                                                                                                                                                                                                                                                                                                                         \\ 
\hline
\multicolumn{6}{c}{\textbf{(a) General Appearance}}                                                                                                                                                                                                                                                                                                                    \\ 
% \hline
blurry                                                    & male                                                        & young                                                     & chubby                                                 & pale\_skin                                                 & rosy\_cheeks                                               \\ 
% \hline  
% \arrayrulecolor[rgb]{0.753,0.753,0.753}\hline

oval\_face                                                & \begin{tabular}[c]{@{}c@{}}receding\\ hairline\end{tabular} & bald                                                      & bangs                                                  & black\_hair                                                & blond\_hair                                                \\ 
% \hline
gray\_hair                                                & brown\_hair                                                 & \begin{tabular}[c]{@{}c@{}}straight\\ hair\end{tabular}   & wavy\_hair                                             & attractive                                                 & \begin{tabular}[c]{@{}c@{}}arched\\ eyebrows\end{tabular}  \\ 
% \hline
\begin{tabular}[c]{@{}c@{}}bushy\\ eyebrows\end{tabular}  & bags\_under\_eyes                                           & eyeglasses                                                & mouth\_slightly\_open                                  & smiling                                                    & big\_nose                                                  \\ 
% \hline
pointy\_nose                                              & \begin{tabular}[c]{@{}c@{}}high\\ cheeks\end{tabular}       & big\_lips                                                 & double\_chin                                           & no\_beard                                                  & 5\_o\_clock shadow                                         \\ 
% \hline
goatee                                                    & sideburns                                                   & mustache                                                  & \begin{tabular}[c]{@{}c@{}}heavy\\ makeup\end{tabular} & \begin{tabular}[c]{@{}c@{}}wearing\\ earrings\end{tabular} & wearing\_hat                                               \\ 
% \hline
\begin{tabular}[c]{@{}c@{}}wearing\\ lipstick\end{tabular} & \begin{tabular}[c]{@{}c@{}}wearing\\ necklace\end{tabular}  & \begin{tabular}[c]{@{}c@{}}wearing\\ necktie\end{tabular} & narrow\_eyes                                           &                                                            &                                                            \\ 
\hline
% \midrule
\multicolumn{6}{c}{\textbf{(b) Detailed Appearance}}                                                                                                                                                                                                                                                                                                                   \\ 
% \hline
Mole                                                      & freckle                                                     & one\_eyed                                                 & scar                                                   & dimple                                                     &                                                            \\ 
% \hline 
\midrule
\multicolumn{6}{c}{\textbf{(c) Light Conditions}}                                                                                                                                                                                                                                                                                                                      \\ 
% \hline
dark                                                      & normal                                                      & bright                                                    & warm white                                             & cool white                                                 & daylight                                                   \\ \hline \midrule
\multicolumn{6}{c}{\textbf{Dynamic Attributes}}                                                                                                                                                                                                                                                                                                                        \\ 
\hline
\multicolumn{6}{c}{\textbf{(a) Action}}                                                                                                                                                                                                                                                                                                                                \\ 
% \hline
blow                                                      & chew                                                        & close\_eyes                                               & cough                                                  & cry                                                        & drink                                                      \\ 
% \hline
eat                                                       & frown                                                       & gaze                                                      & glare                                                  & head\_wagging                                              & kiss                                                       \\ 
% \hline
laugh                                                     & listen\_to\_music                                           & look\_around                                              & make\_a\_face                                          & nod                                                        & play\_instrument                                           \\ 
% \hline
read                                                      & shake\_head                                                 & shout                                                     & sigh                                                   & sing                                                       & sleep                                                      \\ 
% \hline
smile                                                     & smoke                                                       & sneeze                                                    & sniff                                                  & sneer                                                      & talk                                                       \\ 
% \hline
turn                                                      & weep                                                        & whisper                                                   & win                                                    & yawn                                                       & blink                                                      \\ 
% \hline
squint                                                    &                                                             &                                                           &                                                        &                                                            &                                                            \\ 
\hline 
% \midrule
\multicolumn{6}{c}{\textbf{(b) Emotion}}                                                                                                                                                                                                                                                                                                                               \\ 
% \hline
Neutral                                                   & Happy                                                       & Sad                                                       & Anger                                                  & Fear                                                       & Surprise                                                   \\ 
% \hline
Contempt                                                  & Disgust                                                     &                                                           &                                                        &                                                            &                                                            \\ 
\hline 
% \midrule
\multicolumn{6}{c}{\textbf{(c) Light Directions}} \\ 
% \hline
front                                                     & left\_45                                                    & right\_45                                                 & left\_90                                               & right\_90                                                  & back                                                       \\
\bottomrule
\end{tabular}}
\end{table*}

\subsection{Selected Algorithms}
For effective and accurate annotation algorithms, we labeled CelebV-Text using an open-source algorithm\footnote{\url{https://github.com/ewrfcas/face_attribute_classification_pytorch}}.
We follow~\cite{cct2} for light color temperature and we simplify the light intensity calculation by using perceived brightness~\cite{brightness}.
We follow~\cite{emotion} for 8 emotion classification, where emotion label is given for each video frame. We further apply sliding window smoothing algorithm~\cite{savgol} on the temporal domain to smooth the distribution of emotion along time.
All automatically annotated labels are further reviewed by our human annotators.

\section{Template Designs}
For template design, we first employ trained probabilistic natural language English parsers~\cite{de2006generating, nlpparser} to parse the natural language inputs provided by out annotators and get parsing tree banks that appear the most.
Then we modify the parsing to reversely generate descriptions that are near natural languages.
We further choose probabilistic context free grammars (PCFG) to increase the diversity of the generated sentences.
One PCFG template used to generate language descriptions for our general face appearance is shown in Table~\ref{tab:pcfg_general}.
Note that all terminal symbols are bold, and terminal symbol with underlines are dependent on the annotated results.
Specifically, \uline{\textbf{gender\_related\_attributes}} is related the gender, which is a unique value.
\uline{\textbf{personal\_noun}} is also gender related and can be considered as a list where only one single option is picked (\ie, man, woman, male, female).
\uline{\textbf{wear\_related\_attributes}} contains a list of general attributes related to wearing (\ie, heavy makeup, earrings, hat, lipstick, necklace, necktie, eyeglasses).
\uline{\textbf{is\_related\_attributes}} contains a list of general attributes such as bald, young, blurry.
\uline{\textbf{has\_related\_attributes}} contains 5 o'clock shadow, bags under eyes, arched eyebrows, and so on.
% , big lips, big nose, black hair, brown hair, gray hair, long hair, blond hair, bushy eyebrows and so on.
Please refer to our GitHub for all designed templates.
After obtaining the full sentence, we further use NLTK~\cite{nltk} for synonym replacement to increase our generation diversity.
\begin{table}
\centering
\small
\caption{Detailed PCFG design for generating descriptions for general faces.}
\resizebox{0.45\textwidth}{!}{
\begin{tabular}{lll} 
\hline
Rule           &                             & Probability  \\
S              & $\longrightarrow$ NP VP                       & 1.0            \\
NP             & $\longrightarrow$ Det Gender                  & 0.5          \\
NP             & $\longrightarrow$ PN                          & 0.5          \\
VP             & $\longrightarrow$ Wearing PN Are PN HaveWith  & 0.166        \\
VP             & $\longrightarrow$ Wearing PN HaveWith PN Are  & 0.166        \\
VP             & $\longrightarrow$ Are PN HaveWith PN Wearing  & 0.166        \\
VP             & $\longrightarrow$ Are PN Wearing PN HaveWith  & 0.166        \\
VP             & $\longrightarrow$ HaveWith PN Are PN Wearing  & 0.166        \\
VP             & $\longrightarrow$ HaveWith PN Wearing PN Are  & 0.166        \\
Wearing        & $\longrightarrow$ WearVerb WearAttributes     & 1.0            \\
Are            & $\longrightarrow$ IsVerb IsAttributes         & 1.0            \\
HaveWith       & $\longrightarrow$ HaveVerb HaveAttributes     & 1.0            \\
Det            & $\longrightarrow$ \textbf{a}                           & 0.333        \\
Det            & $\longrightarrow$ \textbf{the}                         & 0.333        \\
Det            & $\longrightarrow$ \textbf{this}                        & 0.333        \\
Gender         & $\longrightarrow$ \uline{\textbf{gender\_related\_attributes}} & 0.8          \\
Gender         & $\longrightarrow$ \textbf{person}                      & 0.2          \\
PN             & $\longrightarrow$ \uline{\textbf{personal\_noun}}              & 1.0            \\
WearVerb       & $\longrightarrow$ \textbf{is wearing}                  & 0.5          \\
WearVerb       & $\longrightarrow$ \textbf{wears}                       & 0.5          \\
WearAttributes & $\longrightarrow$ \uline{\textbf{wear\_related\_attributes}}   & 1.0            \\
IsVerb         & $\longrightarrow$ \textbf{is}                          & 1.0            \\
IsAttributes   & $\longrightarrow$ \uline{\textbf{is\_related\_attributes}}     & 1.0            \\
HaveVerb       & $\longrightarrow$ \textbf{has}                         & 0.5          \\
HaveVerb       & $\longrightarrow$ \textbf{has got}                     & 0.5          \\
HaveAttributes & $\longrightarrow$ \uline{\textbf{has\_related\_attributes}}    & 1.0            \\
\hline
\end{tabular}}
\label{tab:pcfg_general}
\end{table}

\section{Results of $n$-grams}
We further compare more unique n-grams among MM-Vox\cite{mmvid}, CelebV-HQ~\cite{celebvhq}, and CelebV-Text in Table~\ref{tbl:ngrams}.
The improvement of our CelebV-Text over MM-Vox\cite{mmvid}, CelebV-HQ~\cite{celebvhq} is quite obvious, which indicates CelebV-Text presents more diverse descriptions.

\begin{table}[t]
\footnotesize
\caption{\textbf{Number of unique $n$-grams.} The numbers of unique $n$-grams for MM-Vox, CelebV-HQ, and CelebV-Text.}
% \vspace{-2mm}
\centering
\resizebox{0.45\textwidth}{!}{
\begin{tabular}{l|cccc}
\toprule
 % & \multicolumn{4}{c}{\#unique \textit{n}-grams} \\ \hline
\multicolumn{1}{l|}{Dataset} & \multicolumn{1}{c}{1-grams} & \multicolumn{1}{c}{2-grams} & \multicolumn{1}{c}{3-grams} & \multicolumn{1}{c}{4-grams} \\ \midrule
MM-Vox~\cite{mmvid} & 65 & 243 & 1478 & 3935 \\
CelebV-HQ~\cite{celebvhq} & 103 & 372 & 1866 & 4932 \\
CelebV-Text & \textbf{593} & \textbf{3385} & \textbf{14,136} & \textbf{45,692} \\ \midrule
\end{tabular}}
% \vspace{-4mm}
\label{tbl:ngrams}
\end{table}

\section{Additional Experiments}
\subsection{FVD/FID/CLIPSIM Settings}
We leverage FVD\footnote{\url{https://github.com/mseitzer/pytorch-fid}}~\cite{fvd}, FID\footnote{\url{https://github.com/sihyun-yu/digan/tree/master/src/metrics}}~\cite{fid}, and CLIPSIM\footnote{\url{https://github.com/openai/CLIP}}~\cite{mmvid} to assess the video temporal consistency, individual frame quality, and relevance between the generated video and input text.
As all metrics are sensitive to data scale during testing, we first randomly select 2,048 videos from the test data as our ``test set'', which are used as the ``real'' part in our metric experiments.
For the facial text-to-video generation task under different training conditions (\eg, trained on CelebV-Text with only general appearance descriptions or with light condition descriptions), 2,048 video samples are also generated from our trained models, which are as used as the ``fake'' part.
To provide enough images for FID testing, 4 frames are uniformly sampled from each video.
In total, we have 8192 images for the real data and fake data respectively.
For both FVD and CLIPSIM evaluation, we follow~\cite{imagen-video} to generate 2048 ``fake'' video samples and compute the metric scores between 2048 real and fake video samples.
For CLIPSIM, we take the average score over all frames.

% quantitative evaluation table
\begin{table}
\caption{\textbf{Benchmark of text-to-video generation on different datasets.} $\downarrow$ means a lower value is better and $\uparrow$ means the opposite.
}
\centering
\begin{subtable}{\linewidth}
\caption{Quantitative results on static descriptions, such as detailed appearance and light conditions descriptions.} \label{tab:ours_supp}
\resizebox{1\linewidth}{!}{
\begin{tabular}{l|l|l|l|l}
\toprule
Dataset & Method & FVD($\downarrow$) & FID($\downarrow$) & CLIPSIM($\uparrow$) \\ \midrule
\multirow{2}{*}{\begin{tabular}[c]{@{}c@{}}CelebV-Text\\ \textbf{Detail App.} \end{tabular}} & TFGAN~\cite{tfgan} & 415.89 $\pm$ 1.11 & 601.46 $\pm$ 15.12 & 0.155 $\pm$ 0.023 \\ 
 & MMVID~\cite{mmvid} & \textbf{68.17 $\pm$ 1.22} & \textbf{58.89 $\pm$ 5.172} & \textbf{0.191 $\pm$ 0.016} \\ \midrule
 \multirow{2}{*}{\begin{tabular}[c]{@{}c@{}}CelebV-Text\\ \textbf{Light Cond.} \end{tabular}} & TFGAN~\cite{tfgan} & 443.95 $\pm$ 2.23 & 591.00 $\pm$ 17.31 & 0.154 $\pm$ 0.020\\ 
 & MMVID~\cite{mmvid} & \textbf{69.41 $\pm$ 2.01} & \textbf{62.88 $\pm$ 4.94} & \textbf{0.187 $\pm$ 0.024} \\ \bottomrule
\end{tabular}}
\end{subtable}

\bigskip
\begin{subtable}{\linewidth}
\caption{Quantitative results on dynamic descriptions of CelebV-Text.}
\centering
\resizebox{1\textwidth}{!}{
\begin{tabular}{l|l|l|l|l}
\toprule
Dataset & Method & FVD($\downarrow$) & FID($\downarrow$) & CLIPSIM($\uparrow$) \\
\midrule
\multirow{3}{*}{\begin{tabular}[c]{@{}l@{}}CelebV-Text\\ \textbf{Light Dir.} \end{tabular}}
& TFGAN~\cite{tfgan} & 433.02 $\pm$ 2.23  & 608.58 $\pm$ 16.93  & 0.156 $\pm$ 0.021  \\
& MMVID~\cite{mmvid} & 69.19 $\pm$ 1.32 & 77.25 $\pm$ 4.05 & 0.172 $\pm$ 0.019 \\ 
& MMVID-interp & \textbf{61.55 $\pm$ 1.28} & \textbf{60.13 $\pm$ 4.17} &  \textbf{0.175 $\pm$ 0.014} \\ 
\midrule
\multirow{3}{*}{\begin{tabular}[c]{@{}l@{}}CelebV-Text\\ \textbf{Emo.+Act.+Light Dir.} \end{tabular}}
& TFGAN~\cite{tfgan} & 597.61 $\pm$ 4.96  & 799.14 $\pm$ 23.66  & 0.148 $\pm$ 0.039 \\
& MMVID~\cite{mmvid} & 118.70 $\pm$ 3.74 & 107.05 $\pm$ 5.48 & 0.171 $\pm$ 0.023  \\ 
& MMVID-interp & \textbf{100.08 $\pm$ 3.48} & \textbf{100.68 $\pm$ 5.21}  & \textbf{0.173 $\pm$ 0.024}  \\ 
\bottomrule
\end{tabular}
}
\end{subtable}
\label{tab:benchmark_supp}
\end{table}

\subsection{Performance Under Texts of Different Lengths}
\begin{figure*}[ht]
    \centering   
    \includegraphics[width=0.97\textwidth]{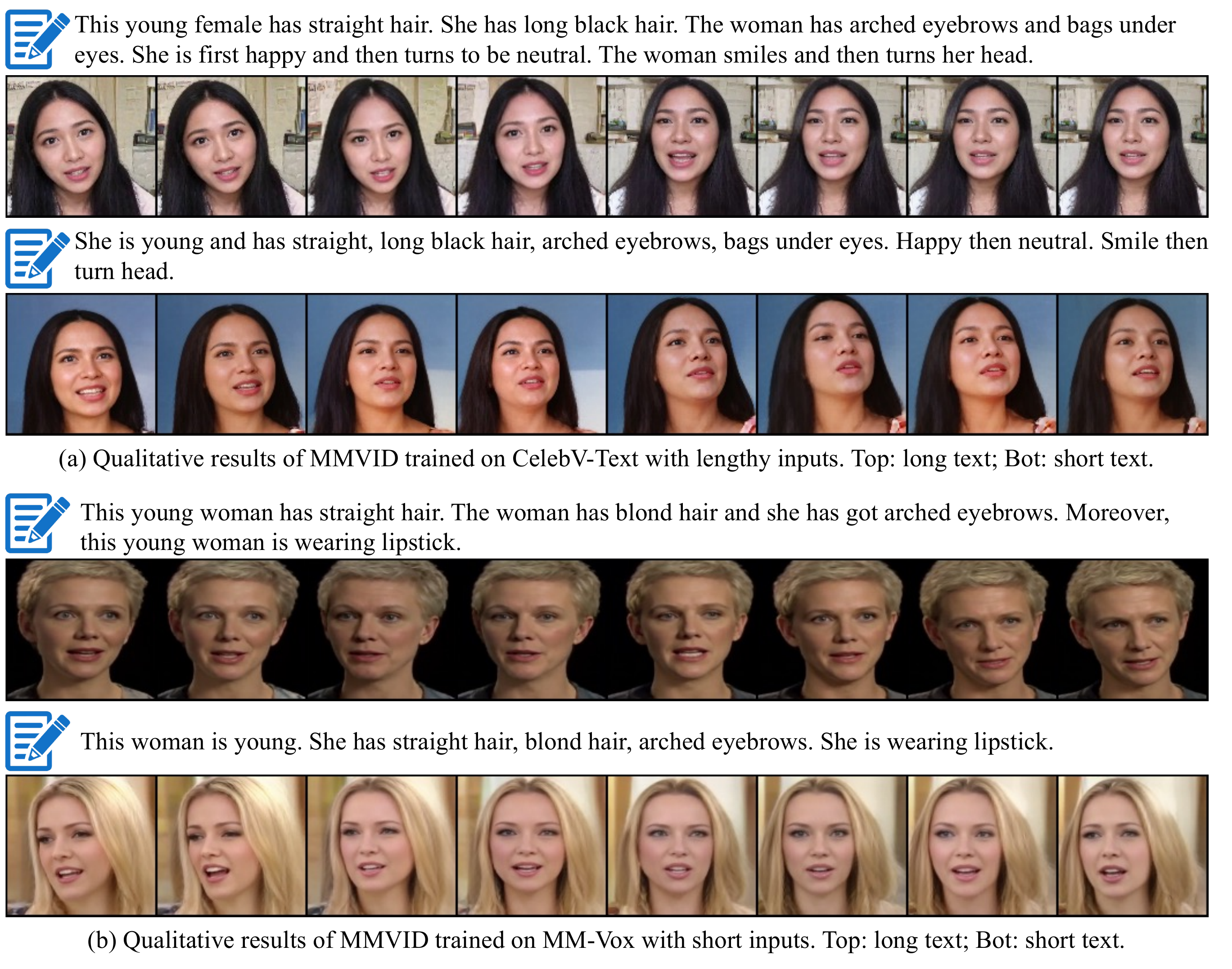}
    \caption{Text-to-video generation with short and lengthy input texts.}
    \label{fig:lengthy_input}
\end{figure*}

We show the model performance trained with text of different lengths while representing the same meaning in Figure~\ref{fig:lengthy_input}.
We discuss that lengthy inputs are closer to the distribution of the natural languages, and it is beneficial to train models with lengthy inputs due to attribute matching.
Specifically, MMVID~\cite{mmvid} trained on CelebV-Text with lengthy inputs produces satisfactory outputs when tested on short texts (Figure~\ref{fig:lengthy_input}~(a)). However, outputs generated by MMVID~\cite{mmvid} trained on MM-Vox~\cite{mmvid} with short texts hardly reflect all attributes given long texts (e.g., straight hair in Figure~\ref{fig:lengthy_input}~(b)).
However, due to the limitation of baseline models, lengthy inputs would reduce the fidelity of output videos (FVD/FID in Table 4 of the main paper), which could be a new direction to devoted.

\subsection{Unconditional Video Generation}
\begin{figure*}
    \centering   
    \includegraphics[width=0.97\textwidth]{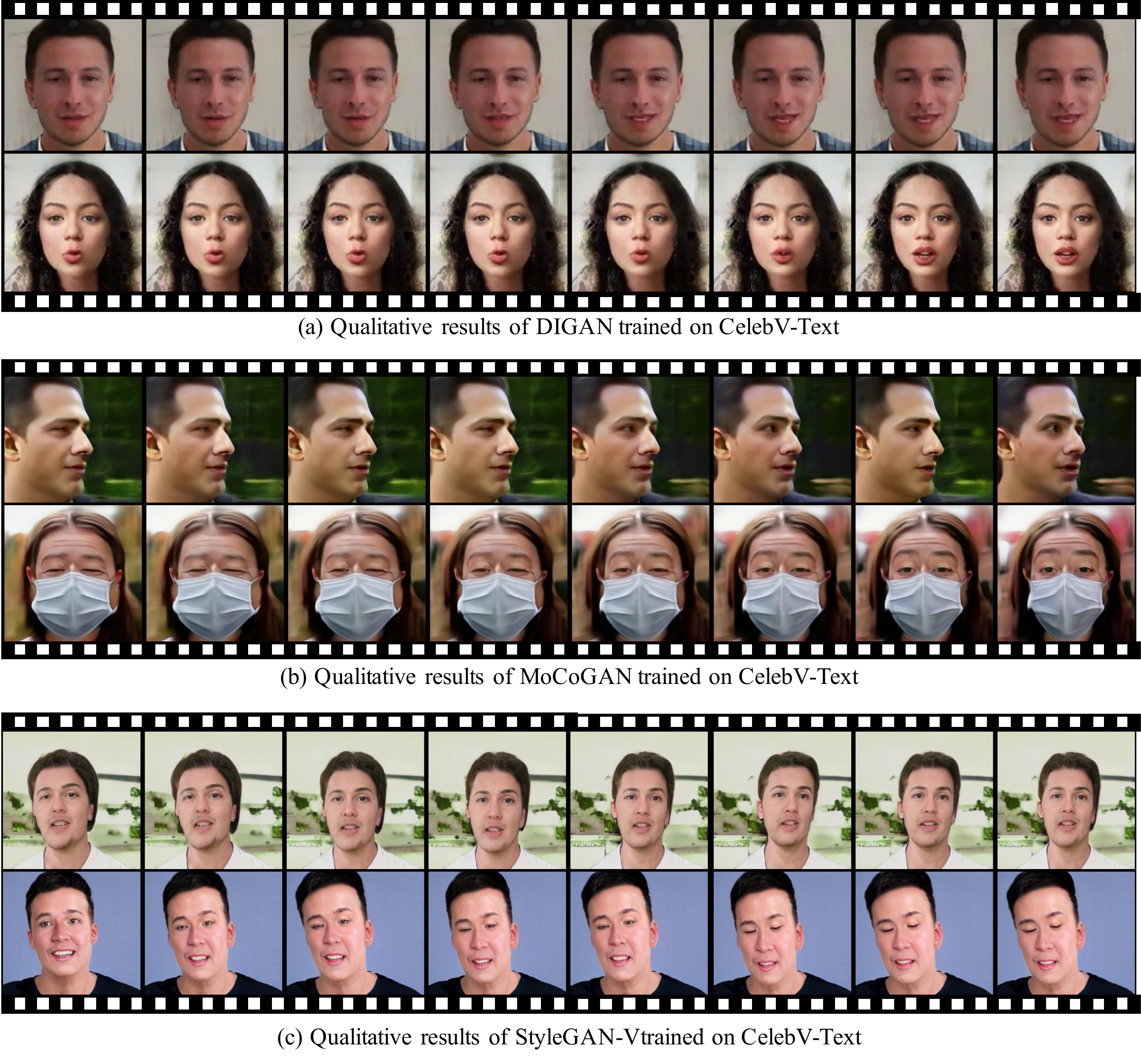}
    \caption{Unconditional video generation results.}
    \label{fig:uncond}
\end{figure*}

To give a more comprehensive and global view of the quality of our dataset, we conduct unconditional video generation with various modern methods (\ie, DIGAN~\cite{digan}, MoCoGAN-HD~\cite{mocogan} and StyleGAN-V~\cite{styleganV}). Results are shown in Figure~\ref{fig:uncond}.

\subsection{Static Face Video Generation}
To further demonstrate the practical effectiveness of our CelebV-Text for facial text-video generation tasks, we additional present our generation results both quantitatively and qualitatively.
As shown in Table~\ref{tab:benchmark_supp}~(a), results of TFGAN~\cite{tfgan} and MMVID~\cite{mmvid} trained on both CelebV-Text with text descriptions about detailed appearances and light conditions are listed.
We can see that MMVID~\cite{mmvid} performs better than TFGAN~\cite{tfgan} under both conditions.

\begin{figure*}
    \centering   
    \begin{subfigure}[b]{\textwidth}
         \centering
         \includegraphics[width=0.97\textwidth]{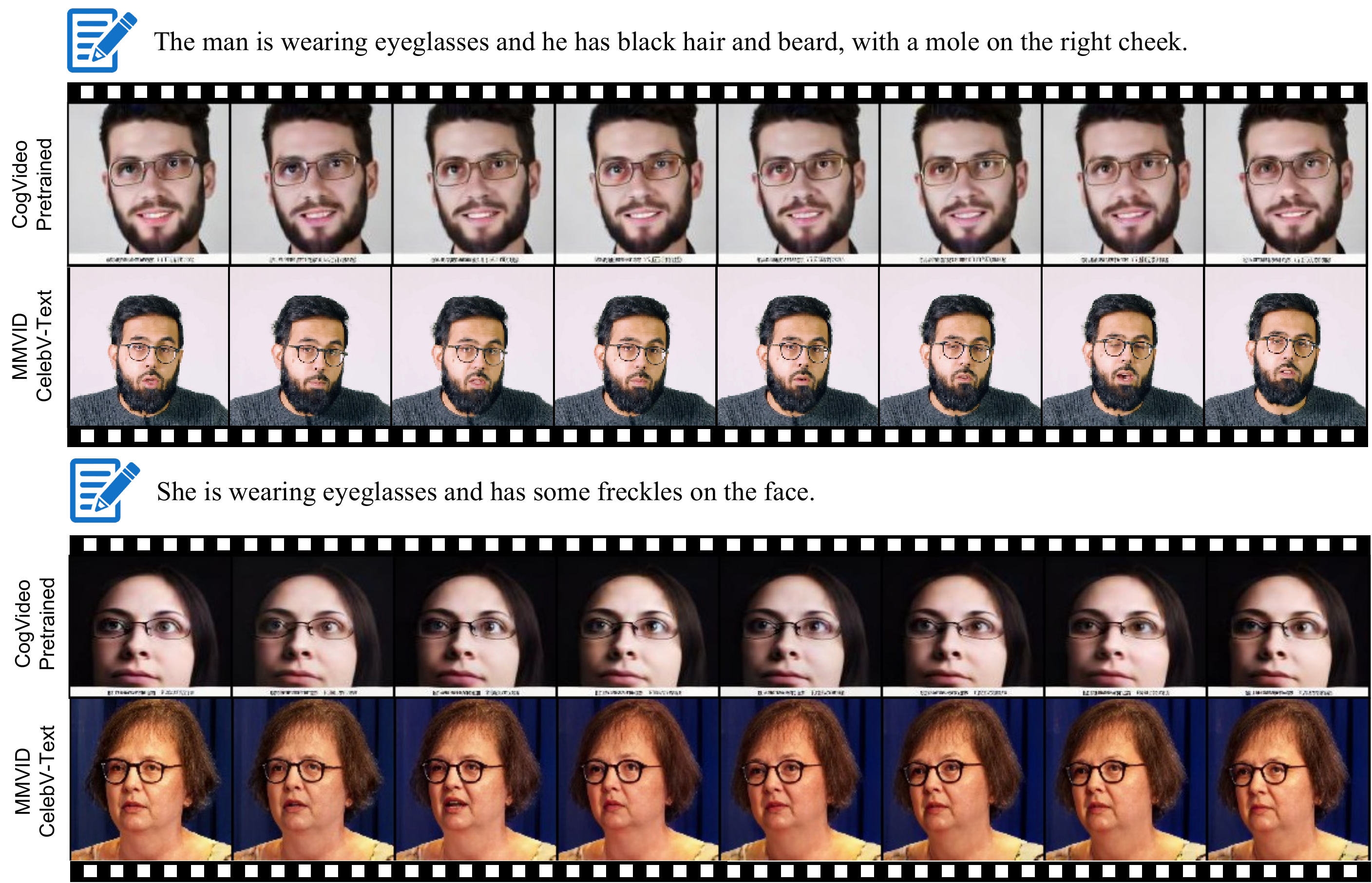}
          \caption{Static - Detailed Appearance}
         \label{fig:face_details}
     \end{subfigure}
     \begin{subfigure}[b]{\textwidth}
         \centering
         \includegraphics[width=0.97\textwidth]{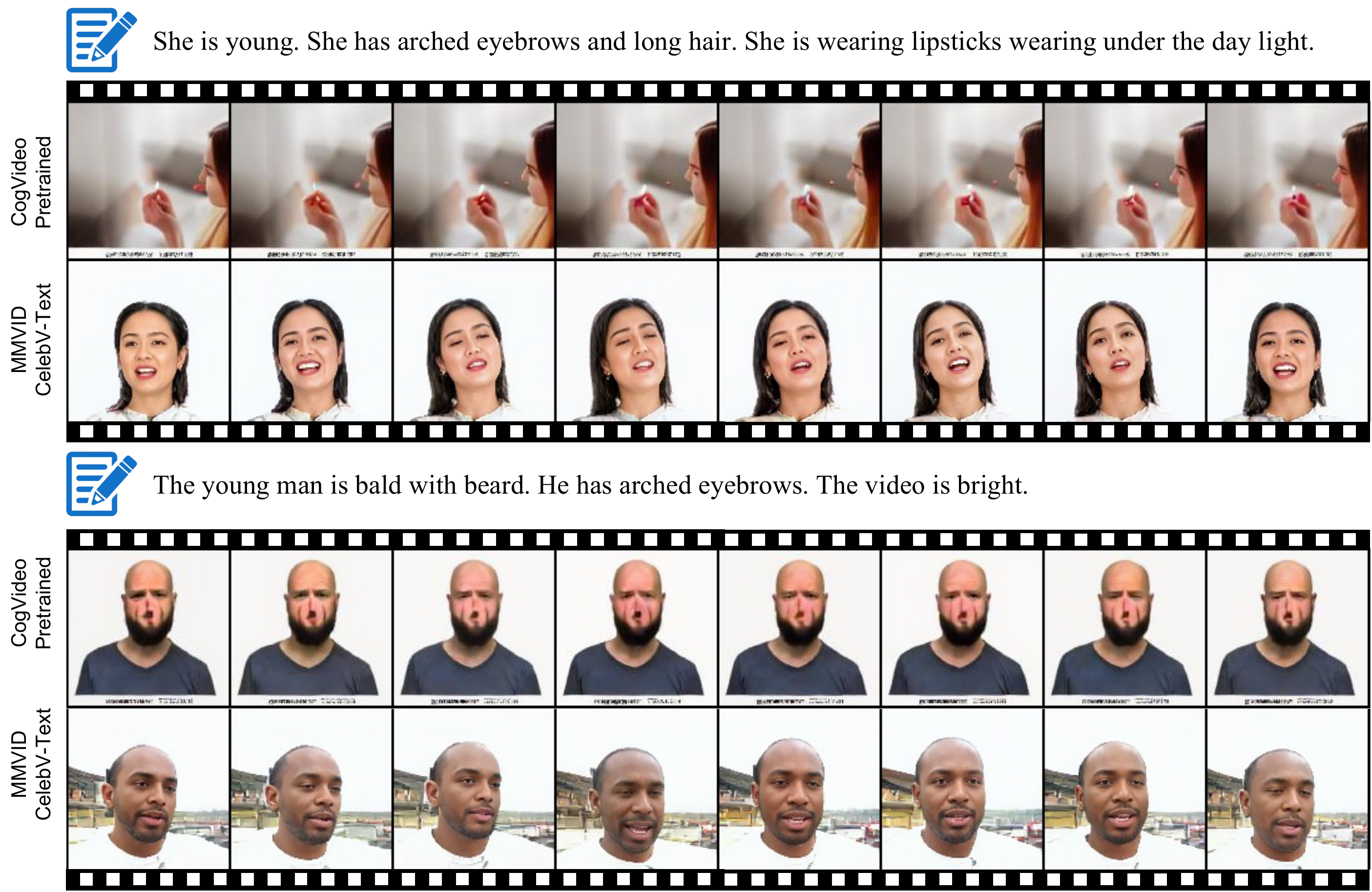}
          \caption{Static - Light Conditions}
         \label{fig:light_condition}
     \end{subfigure}
     \vspace{-7mm}
    \caption{\textbf{Qualitative results of facial text-to-video generation on static descriptions.} The video samples are generated given texts describing static (a) detailed appearance and (b) light conditions.} \label{fig:static}
\end{figure*}

\begin{figure*}
    \centering   
    \includegraphics[width=0.97\textwidth]{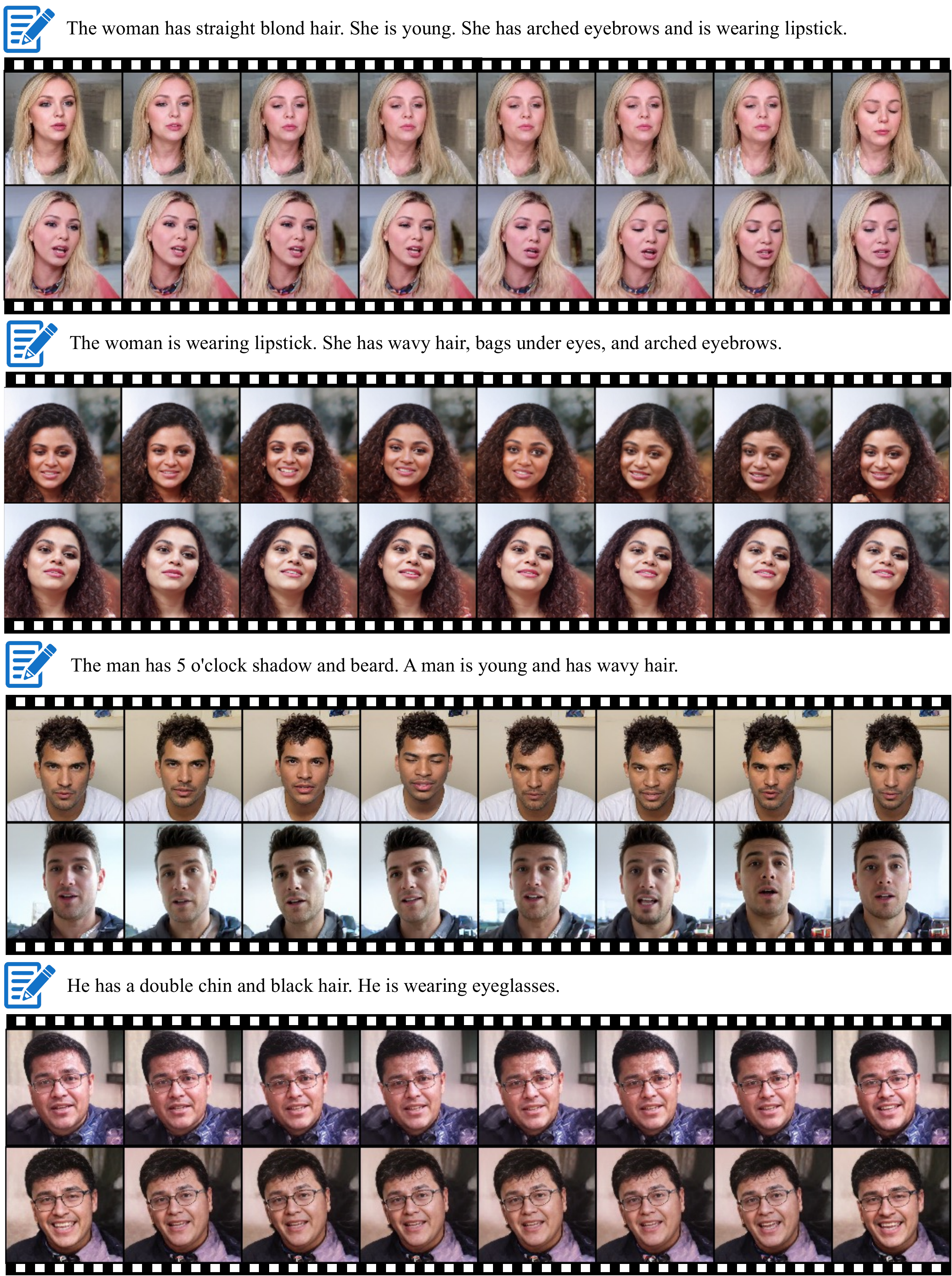}
    \vspace{-2mm}
    \caption{More sampled results from MMVID with input texts describing general appearances.}
    \label{fig:general}
\end{figure*}

In addition, we also compare the model performance of MMVID~\cite{mmvid} with CogVideo~\cite{cogvideo}.
To validate the effectiveness of our facial text-video dataset in static attributes, we show more visualization samples in Figure~\ref{fig:static} trained on CelebV-Text with the descriptions of static attributes (\ie, detailed appearance and light conditions).
We can see that although CogVideo~\cite{cogvideo} is trained on large-scale text-video dataset with larger model size than MMVID~\cite{mmvid}, MMVID~\cite{mmvid} trained on CelebV-Text can give much better results where the generated video samples correspond well with the text input.
More results by MMVID~\cite{mmvid} trained on general appearance are shown in Figure~\ref{fig:general}.
These results validate the effectiveness of our CelebV-Text.

% for StarGAN-v2
% More quantitative results of both TFGAN~\cite{tfgan} and MMVID~\cite{mmvid} trained on CelebV-Text with text descriptions about detailed appearances and light conditions are shown in Table~\ref{tab:benchmark}.
% Note that both CelebV-Text with detailed appearances and CelebV-Text with light conditions include text descriptions about general appearance attributes as well, which is omitted here for simplified representation.
% We can see that 

\subsection{Dynamic Face Video Generation}
We show more quantitative and qualitative results when text descriptions about dynamic attributes are used for training.
For all experiments, we report results of MMVID~\cite{mmvid}, MMVID-interp~\cite{mmvid}, and CogVideo~\cite{cogvideo} both quantitatively and qualitatively.

\begin{figure*}
\centering
\includegraphics[width=0.97\textwidth]{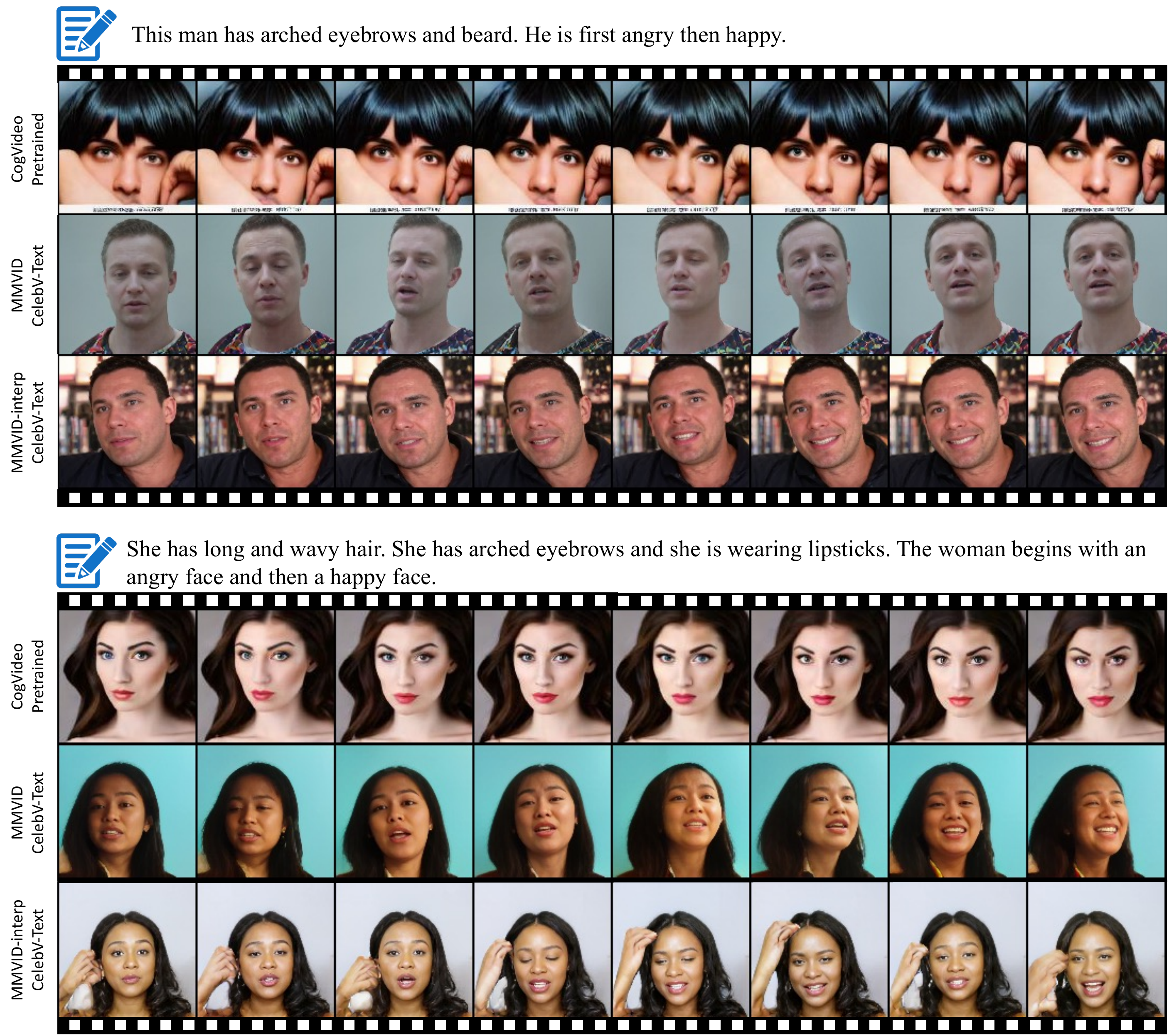}
\caption{\textbf{Qualitative results of facial text-to-video generation.} The video samples are generated given texts describing dynamic emotion.}
\label{fig:face_dynamic_emotion}
\end{figure*}

\begin{figure*}
\centering
\includegraphics[width=0.97\textwidth]{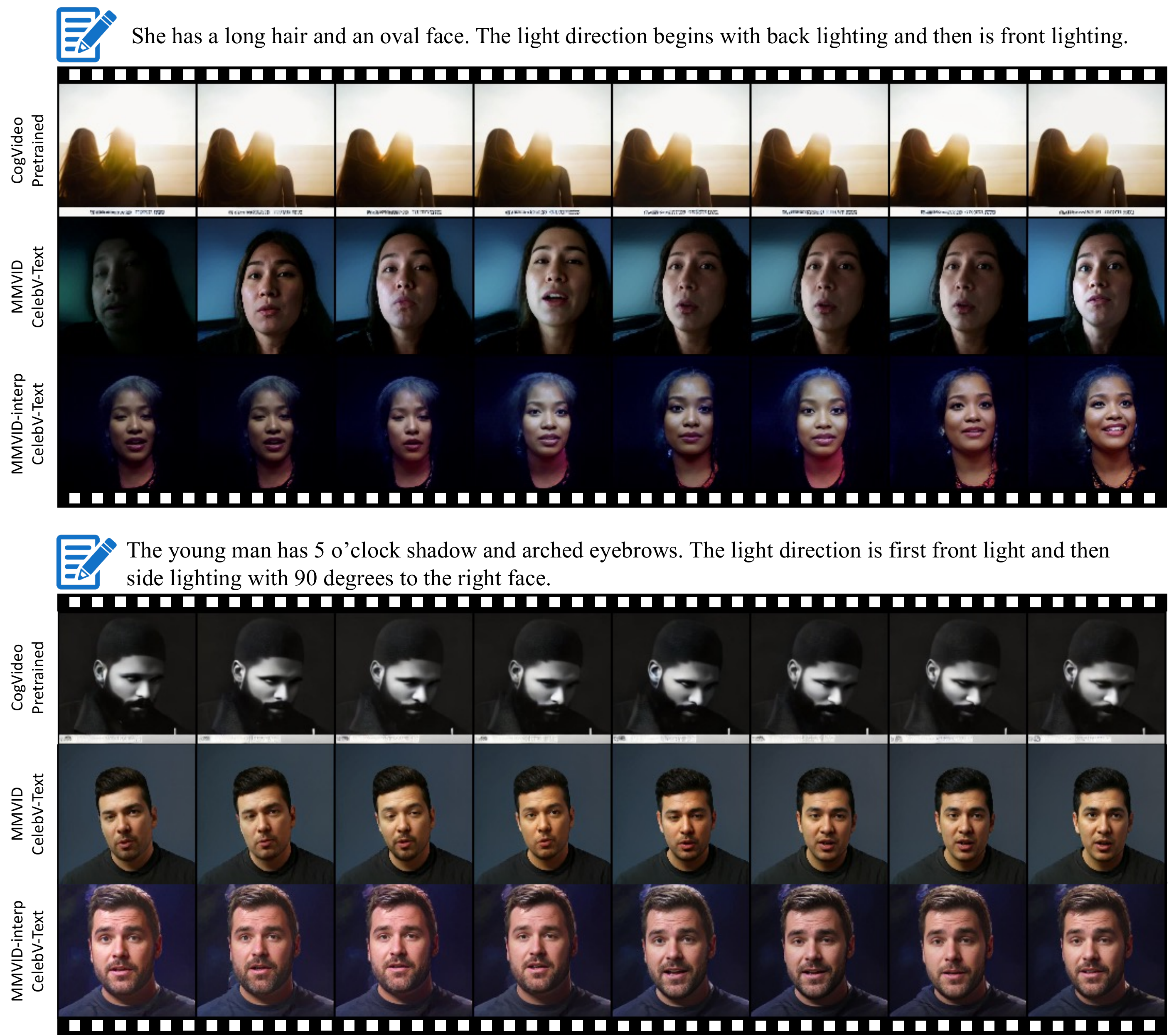}
\caption{\textbf{Qualitative results of facial text-to-video generation on dynamic descriptions.} The video samples are generated given texts describing dynamic light directions.}
\label{fig:face_dynamic_light}
\end{figure*}

We report more quantitative results of CelebV-Text with variant input texts in Table~\ref{tab:benchmark_supp}~(b) and qualitative results of dynamic emotion and light direction changes in Figure~\ref{fig:face_dynamic_emotion} and Figure~\ref{fig:face_dynamic_light}, respectively.

\noindent{\textbf{MMVID-interp.}}
As mentioned in the main work, we follow~\cite{tfgan} to apply test-time interpolation to MMVID~\cite{mmvid} to improve text encoding and better understand the dynamics.
Specifically, given the text input describing dynamic attribute changes, we manually split the dynamic description into two sentences, \ie, $S_{1}$ and $S_{2}$.
$S_{1}$ contains the description about the appearance and the first dynamic attribute, and $S_{2}$ contains the description about the appearance and the second dynamic attribute.
Let $\textbf{t}_{S_{1}}$ and $\textbf{t}_{S_{2}}$ denote the feature representation obtained from the text encoder used in MMVID~\cite{roberta}.
In this case, the description about appearance is repeated twice, so that the text encoding of it can be emphasized and improved, making the generation process more stable on preserving face identities.
During the sampling process, the encoded text condition $\textbf{t}$ is obtained by a linear interpolation between $\textbf{t}_{S_{1}}$ and $\textbf{t}_{S_{2}}$:
\begin{equation}
    \textbf{t}_{i} = (1-\alpha_{i})\textbf{t}_{S_{1}} + \alpha_{i} \textbf{t}_{S_{2}},
\end{equation}
where $\alpha_{i}$ is proportional to the text sequence length.
Our modification is simple and will be improved in the future.

\clearpage
\FloatBarrier

\end{document}